\documentclass[11pt, a4paper, twocolumn, confidential, copyright, google]{google}
\pdfmapfile{+samsung.map}

\usepackage[authoryear, sort&compress, round]{natbib}
\usepackage{wrapfig}
\usepackage{adjustbox}
\usepackage{hyperref}
\usepackage{url}
\usepackage{graphicx}
\usepackage{amsmath}
\usepackage{amssymb}
\usepackage{booktabs}
\usepackage{wrapfig}
\usepackage{pifont}
\usepackage{colortbl}
\usepackage{xcolor}
\usepackage{adjustbox}
\usepackage{siunitx}
\usepackage{multirow}
\usepackage{tocloft}
\definecolor{lightgray}{gray}{0.9}
\definecolor{rowgray}{gray}{0.95}
\usepackage{subcaption}


\uselogo{} 

\title{HiMAE: Hierarchical Masked Autoencoders Discover Resolution-Specific Structure in Wearable Time Series}

\correspondingauthor{simonlee711@g.ucla.edu/s.lee5@partner.samsung.com}

\reportnumber{} 

\author[*,1,2]{Simon A. Lee}
\author[1]{Cyrus Tanade}
\author[1]{Hao Zhou}
\author[1]{Juhyeon Lee}
\author[1]{Megha Thukral}
\author[1]{Minji Han}
\author[1]{Rachel Choi}
\author[1]{Md Sazzad Hissain Khan}
\author[1]{Baiying Lu}
\author[1]{Migyeong Gwak }
\author[1]{Mehrab Bin Morshed}
\author[1]{Viswam Nathan}
\author[1]{Md Mahbubur Rahman}
\author[1]{Li Zhu}
\author[1]{Subramaniam Venkatraman }
\author[1]{Sharanya Arcot Desai }

\affil[*]{Work done during AI Residency}
\affil[1]{Digital Health Team, Samsung Research America}
\affil[2]{Department of Computational Medicine, University of California Los, Angeles}

\begin{abstract}
Wearable sensors provide abundant physiological time series, yet the principles governing their predictive utility remain unclear. We hypothesize that temporal resolution is a fundamental axis of representation learning, with different clinical and behavioral outcomes relying on structure at distinct scales. To test this \emph{resolution hypothesis}, we introduce \emph{HiMAE} (Hierarchical Masked Autoencoder), a self-supervised framework that combines masked autoencoding with a hierarchical convolutional encoder–decoder. HiMAE produces multi-resolution embeddings that enable systematic evaluation of which temporal scales carry predictive signal, transforming resolution from a  hyperparameter into a probe for interpretability. Across classification, regression, and generative benchmarks, HiMAE consistently outperforms state-of-the-art foundation models that collapse scale, while being orders of magnitude smaller. HiMAE is an efficient representation learner compact enough to run entirely on-watch, achieving sub-millisecond inference on smartwatch-class CPUs for true edge inference. Together, these contributions position HiMAE as both an efficient self supervised learning method and a discovery tool for scale-sensitive structure in wearable health.
\end{abstract}

\begin{document}

\maketitle
\onecolumn

\section{Introduction}
\label{intro}
Wearable sensors have emerged as a primary modality for continuous health monitoring, providing access to rich physiological and behavioral signals in free-living settings \citep{erturk2025beyond}. Despite their ubiquity, the utility of wearable signals for machine learning in healthcare remains poorly understood. Unlike images~\citep{dosovitskiy2021imageworth16x16words, simonyan2014deepinsideconvolutionalnetworks, zhou2015learningdeepfeaturesdiscriminative, petsiuk2018riserandomizedinputsampling} or text~\citep{brown2020language, li-etal-2016-visualizing, sundararajan2017axiomatic, arras2017explainingrecurrentneuralnetwork}, physiological time series rarely admit obvious visual cues that map cleanly to clinical outcomes, leaving open fundamental questions about which features carry predictive value. A particularly unresolved issue concerns temporal resolution: should models operate at a single universal resolution, or do different health outcomes depend on resolution-specific structure? Clinically actionable events can arise on second-level timescales, requiring representations that both capture fine-grained temporal patterns and support real-time inference under the computational constraints of wearable devices. We hypothesize that resolution is not a nuisance parameter but a fundamental axis of physiological representation learning. We refer to this as the \textit{resolution hypothesis}, which posits that temporal granularity governs predictive performance in clinical and behavioral tasks. In this framing, “resolution” denotes the effective temporal context over which representations are formed—from fine-scale waveform morphology to coarse-scale dynamics spanning the whole sequence.

From an algorithmic perspective, much of the field defaults to transformer-based architectures~\citep{vaswani2017attention}, implicitly assuming that flexibility and capacity outweigh inductive bias. Yet wearable signals, while long in sequence length, are often generated by a few latent processes driven by biological mechanisms and captured through only a handful of sensor modalities. In this sense they are low-dimensional and highly structured. This raises the possibility that transformers may not only overfit but also obscure resolution-specific structure, rather than expose it. By contrast, hierarchical convolutional biases offer a natural mechanism for aligning architectures with the resolution hypothesis, capturing both local detail and long-range dependencies in a principled way. This motivates a re-examination of architectural design choices for self-supervised learning (SSL) on physiological time series.

\begin{figure}[t!]
    \centering
    \includegraphics[width=\linewidth]{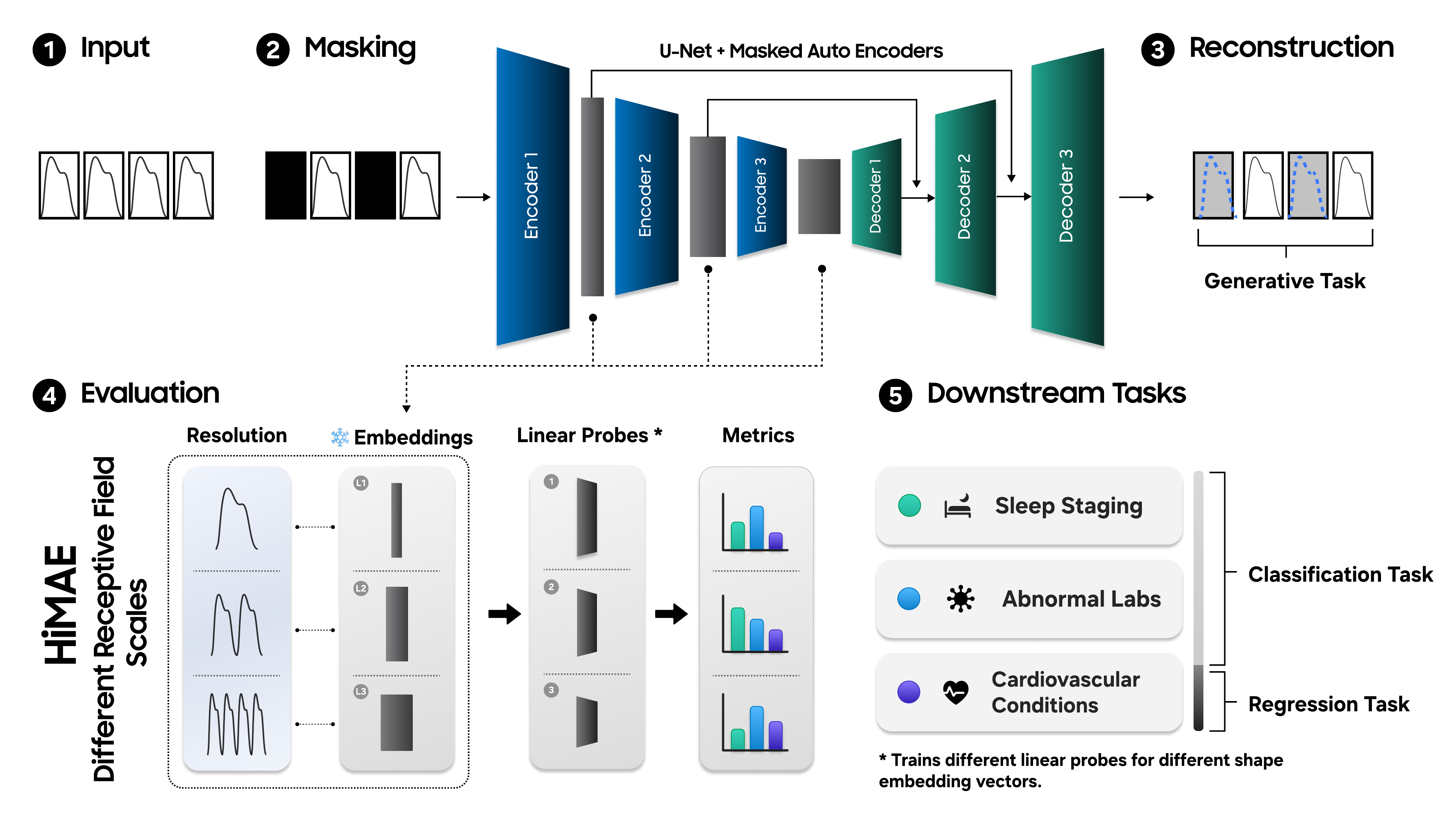}
    \caption{\textbf{HiMAE pre-training and evaluation pipeline.}
(1) Physiological sequences are split into temporal patches.
(2) Selected patches are masked randomly or contiguously.
(3) A U-Net–style CNN encoder–decoder reconstructs missing values, with loss applied only to masked regions.
(4) Multi-resolution embeddings feed linear probes for classification and regression benchmarking.
(5) Three categorized task-lists are evaluated.}
    \label{mainfig}
\end{figure}

In this work, we address these challenges by introducing \textit{HiMAE} (Hierarchical Masked Autoencoder), a self-supervised pretraining framework for wearable time series that directly operationalizes the resolution hypothesis (Figure \ref{mainfig}). HiMAE combines the masked autoencoding paradigm with 1D physiological signals by coupling patch-masking objectives~\citep{wang2023hardpatchesminingmasked} with a U-Net–inspired encoder–decoder~\citep{ronneberger2015u}. Crucially, HiMAE produces multi-resolution embeddings, with each level of the hierarchy corresponding to a distinct temporal granularity. This design enables systematic interrogation of which resolutions carry predictive signal, while simultaneously yielding lightweight, efficient representations. Beyond its architectural advantages, HiMAE allows us to benchmark the resolution hypothesis across 14 classification and regression tasks. Our results reveal resolution-specific structure in wearable signals that is not readily identifiable by human experts, offering new insights into both representation learning and the interpretability of physiological time series in the time domain. Our contributions are threefold:
\begin{itemize}
    \item We introduce \textit{HiMAE}, a hierarchical, scalable, and computationally efficient self supervised learning framework that achieves state-of-the-art performance across generative, classification, and regression benchmarks.
    \item We leverage HiMAE’s (U-Net's) multi-resolution embeddings to probe how temporal scales affect different downstream tasks, leading to discoveries about human physiology.
    \item HiMAE’s compactness enables on-device inference (on smartwatches), which to our knowledge is the first of its kind.
\end{itemize}

\section{Related Work}
\label{related}

\subsection{Self-Supervised Pretraining Objectives for Wearable Signals}

Wearable devices equipped with photoplethysmography (PPG), electrocardiography (ECG), and accelerometry generate long, multi-channel time series encoding diverse physiological and behavioral phenomena, including cardiovascular dynamics~\citep{castaneda2018review}, activity patterns~\citep{yuan2024self, xu2025lsm}, sleep cycles~\citep{li2021transfer, thapa2024sleepfm, logacjov2025long}, and other latent processes. These data streams are abundant, and predominantly unlabeled, making them well suited for large-scale self-supervised learning~\citep{kaplan2020scaling, bommasani2021opportunities, zhou2024comprehensive, liang2024foundation}.

SSL has become the dominant paradigm for wearable time-series representation learning, given the scarcity of labeled data and the ubiquity of unlabeled signals in free-living settings \citep{lee2025foundation}. Among SSL strategies, masked autoencoding has emerged as a central approach, inspired by its success in vision~\citep{he2022masked, vaid2023foundational} and language modeling~\citep{devlin-etal-2019-bert}. The method randomly occludes patches of the signal and tasks the model with reconstructing them, encouraging representations that capture latent physiological structure and temporal regularities~\citep{zhang2022mask, kong2023understanding}. Recent large-scale efforts, most notably Google’s LSM series~\citep{narayanswamy2024scaling, xu2025lsm}, rely heavily on masked autoencoding, establishing it as a pretraining standard for multi-modal wearable datasets. Yet despite its effectiveness for local pattern recovery, vanilla masked autoencoding often struggles to capture multi-resolution features unless coupled with explicitly hierarchical architectures.

In parallel, contrastive learning enforces invariance by pulling semantically similar samples together in latent space while pushing dissimilar ones apart~\citep{schmitt2008measurement, jaiswal2020survey}. The central challenge for wearables is defining positive and negative pairs without labels. One solution is participant-level contrastive training, where samples from the same individual are positives and samples from different individuals are negatives, an approach adopted in Apple’s ECG and PPG foundation models~\citep{abbaspourazad2023large} and closely related to the SimCLR framework~\citep{chen2020simpleframeworkcontrastivelearning}. Other domain-specific innovations define pairs through physiological priors: PaPaGei leverages PPG morphology~\citep{pillai2024papagei}, while SleepFM extends the paradigm across EEG, ECG, and EMG to enforce cross-modal consistency~\citep{thapa2024sleepfm}. Additional embedding-level regularizers, such as differential entropy constraints~\citep{jing2021dimcollapse, abbaspourazad2023large}, further enrich learned representations. However, contrastive methods are highly sensitive to augmentation heuristics (which are rarely physilogically meaningful), computationally intensive, and limited in interpretability, providing little insight into which temporal structures are preserved.

HiMAE departs from both flat masked and contrastive approaches in two ways. First, instead of relying on a single-scale reconstruction or augmentation heuristics, HiMAE couples masked autoencoding with a hierarchical encoder–decoder that integrates information across resolutions, treating temporal scale as an explicit dimension of representation. Second, by extracting embeddings at multiple scales and probing them independently, HiMAE transforms SSL from a pretraining mechanism into a discovery tool: it directly tests which temporal resolutions carry predictive signal for downstream tasks. In doing so, HiMAE preserves the efficiency of masked autoencoding while introducing interpretability absent in contrastive or flat masked objectives.

\subsection{Multi-scale Learning} 

The emphasis on resolution awareness connects naturally to multi-scale learning, where modeling temporal signals across multiple granularities has emerged as a powerful inductive bias. In vision, multi-scale architectures such as pyramidal CNNs and hierarchical attention enable models to integrate fine-scale edges with coarse semantic structures, substantially improving recognition and generation in 2D~\citep{wang2016temporal, yang2016hierarchical, liu2021swin, kusupati2024matryoshkarepresentationlearning, liu2024vision} and 3D~\citep{he2017multi, ghadai2019multi, zhang2022point}.

In time series, multi-scale methods are fewer but increasingly influential. N-HiTS~\citep{challu2022nhitsneuralhierarchicalinterpolation} improves long-horizon forecasting by allocating capacity across frequencies via hierarchical interpolation. Pyraformer~\citep{liu2022pyraformer} leverages pyramidal attention to capture dependencies over a tree of scales, while Scaleformer~\citep{shabani2023scaleformer} introduces iterative refinement across resolutions. Pathformer~\citep{chen2024pathformer} further adapts pathways dynamically to match input-specific temporal dynamics. Together, these approaches highlight that temporal signals are inherently hierarchical and that resolution carries predictive structure rather than being a nuisance variable.

Prior multi-scale methods typically rely on fixed hierarchies or task-specific refinement stages (e.g., for forecasting), which constrains their generality. While HiMAE also inherits inductive biases from convolutional design choices (e.g., step size, padding, kernel width), these parameters define receptive fields rather than dictate which scales are salient. By coupling self-supervised reconstruction with these fields, HiMAE induces a hierarchy of temporal embeddings that can be probed independently. 

\section{Methods}

\subsection{Hierarchical Masked Autoencoders (HiMAE)}

HiMAE combines masked autoencoding \citep{baldi2012autoencoders, he2022masked} with 1-D physiological time series by coupling a patch-masking objective with a U-Net–style convolutional encoder–decoder \citep{ronneberger2015u}. Given an input sequence $x \in \mathbb{R}^{C \times L}$, we partition it into $N = L/P$ non-overlapping patches of length $P$. A binary mask $m \in {0,1}^N$ is sampled from a Bernoulli distribution with parameter $r$, indicating the masking ratio. Masked indices are selected uniformly at random without replacement, expanded to match temporal resolution as $m' \in {0,1}^L$, and applied to the sequence, yielding $\tilde{x} = x \odot (1 - m')$. This masking procedure removes substantial context, forcing the model to infer higher-order dependencies. In addition to random masking, we also employ contiguous masking, in which adjacent patches are removed to mimic sensor dropout similar to recent protocols showing benefits \citep{xu2025lsm}. Both regimes are interleaved during pretraining to promote robustness across reconstruction settings.

The encoder $f_\theta$ is a hierarchical 1D CNN composed of residual convolutional blocks with stride-2 convolutions that downsample the temporal resolution by half at each stage, expanding the receptive field so that deeper layers capture long-range dependencies while shallow layers retain local detail. Each residual block consists of two convolutions with kernel size 5, batch normalization \citep{ioffe2015batchnormalizationacceleratingdeep}, and GELU activations \citep{hendrycks2023gaussianerrorlinearunits}, along with a projection shortcut when input and output dimensions differ. The decoder $g_\phi$ mirrors this structure with transposed convolutions for upsampling and incorporates skip connections from encoder layers, concatenating intermediate features to restore fine-grained temporal structure. All convolutions are standard 1D operations defined over temporal windows, and striding handles subsampling directly. Intermediate activations use GELU, while the final layer applies a $\tanh$ nonlinearity so that outputs $\hat{x} \in \mathbb{R}^{C \times L}$ are bounded in $[-1,1]$, matching the normalized input range.

\begin{wrapfigure}{r}{0.6\textwidth}
  \centering
      \includegraphics[width=0.59\textwidth]{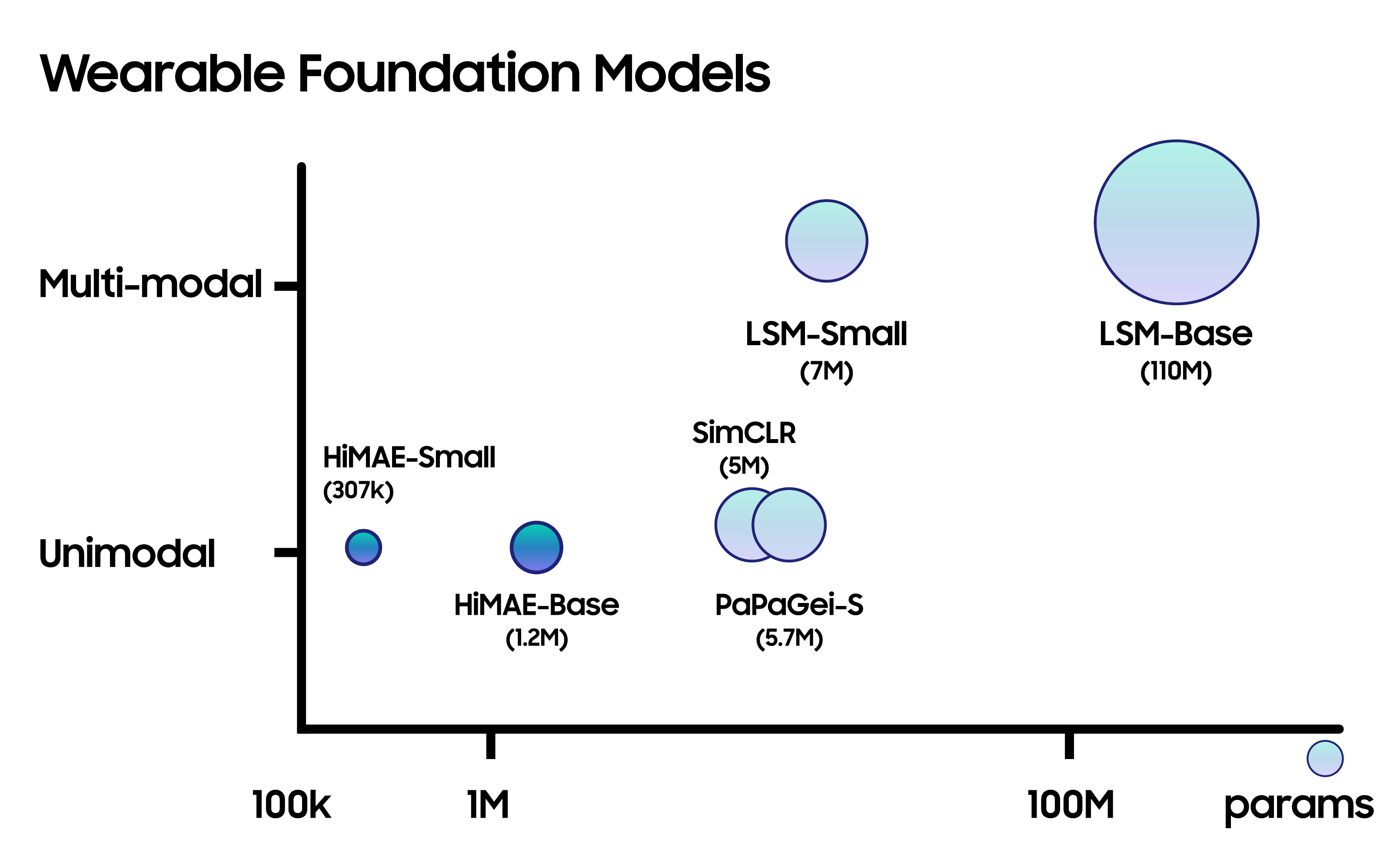}
  \caption{\textbf{HiMAE is lightweight compared to other methods proposed in the literature.}}
  
  \label{size}
\end{wrapfigure}

We deliberately adopt a convolutional U-Net backbone rather than a transformer-based encoder for two reasons. First, physiological signals exhibit strong local dependencies governed by morphology (e.g., PPG waveform shape, ECG peaks), which are naturally modeled by finite receptive fields. Convolutions \citep{oshea2015introductionconvolutionalneuralnetworks} encode this locality directly, whereas transformers must simulate it through restricted attention, often at higher parameter cost. Second, multi-resolution structure is intrinsic to physiology (e.g., heartbeats unfold over milliseconds, rhythms span seconds). A hierarchical CNN with skip connections provides an architectural bias toward such nested timescales, aligning directly with the resolution hypothesis and being orders of magnitude smaller than other proposed foundation models in this space (See Figure~\ref{size} for comparison). In contrast, transformers emphasize global mixing, which may obscure resolution-specific structure while consuming substantially more compute (Table \ref{watch}). This rationale motivates HiMAE’s design as not only efficient but also inductively aligned with the temporal statistics of wearable signals.

Multi-resolution embeddings extracted from different levels of the hierarchy are probed independently, with distinct linear classifiers trained per resolution \citep{alain2018understandingintermediatelayersusing}. This design enables us to systematically evaluate which temporal granularity carries predictive signal for downstream tasks, rather than collapsing embeddings into a single latent space. Finally, choices of patch length $P$ and kernel size were guided by ablations (Appendix Section~\ref{ablation-hyper}), which confirmed that $P=5$ and kernel size 5 yield the best balance between local fidelity and receptive field expansion when all other hyperparameters were fixed.

Training minimizes a masked reconstruction loss restricted to occluded regions: $
\mathcal{L}_{\text{MSE}}(\theta, \phi) = \frac{\| (\hat{x} - x) \odot m' \|_2^2}{\sum_{t=1}^L m'_t},
$ where $m'$ ensures that gradients are only computed on masked segments. This objective estimates $p(x_{\mathcal{M}} | x_{\mathcal{O}})$, with $\mathcal{M}$ and $\mathcal{O}$ denoting masked and observed indices, preventing trivial copying of visible inputs and promoting temporally coherent, multi-scale representations.

\subsection{Pretraining and Evaluation Protocol}

PPG Sequences were sampled at $f_s = 100$ Hz over fixed windows of $T=10$s ($L=1000$ timesteps). 10 second windows were selected due to clinically actionable events occurring in these time scales (ECG is collected at 10s intervals in clinical settings \citep{shuai201610, elgendi2012analysis}) and due to our interest in real-time monitoring on edge devices. Each signal was divided into non-overlapping patches of length $P=5$ (200 patches total), and a masking ratio $r=0.8$ was applied with patterns resampled per sequence and iteration to mitigate overfitting (we empirically tested this masking ratio in Appendix Section \ref{ablation-hyper} with similar observations made in \citep{narayanswamy2024scaling}). The encoder architecture employed channel widths $[16, 32, 64, 128]$, mirrored in the decoder. Optimization was performed with AdamW \citep{loshchilov2019decoupledweightdecayregularization} ($\text{lr}=10^{-3}$, weight decay $=10^{-3}$) using a warmup–cosine schedule (10\% linear warmup steps followed by cosine decay). Models trained up to 100k steps with batch size 2048 and early stopping triggered after 3 epochs without improvement similar to the protocols found in \citep{narayanswamyscaling}. Data splits followed a 90/10 (train/validation) protocol across subjects, ensuring no identity overlap between pretraining and validation. Pretraining converged within 12 hours when distributing training across 4 Tesla T4 GPUs using PyTorch lightning \citep{paszke2019pytorchimperativestylehighperformance}.

\subsection{Pretraining Datasets} We construct our pretraining corpus from approximately $80{,}000$ hours of wearable green PPG signals, drawn from seven large-scale free world studies conducted at Samsung Research and their subsidary branches. These datasets include recordings from $47{,}644$ participants across seven distinct wearable devices, capturing broad demographic, behavioral, and hardware variability in a noisy environment (See Appendix Section \ref{ethics} for ethics considerations). Although our modeling framework is modality-agnostic and can extend to other physiological signals such as electrocardiograms (see Appendix \ref{ecg-fm}), we focus here on PPG due to its prevalence and the scale of available data (we lack the same order of magnitude of ECG compared to PPG because ECG is not passively collected). To ensure reliability, we apply a standardized preprocessing pipeline that retains only high-quality segments, filtering by a Signal Quality Index (SQI). The retained signals are further refined using a bandpass filter of $0.5$–$8$ Hz \citep{christiano2003band}, consistent across all pretraining and evaluation studies, to isolate physiologically relevant dynamics. Finally, signals are normalized to the range $[-1, 1]$ to match the output range of the $\tanh$ activation function used in our models.

\section{Experimental Design}
\label{ed}

We follow the evaluation protocol of \citet{narayanswamy2024scaling} and extend it into a unified benchmark suite spanning generative, classification (and regression tasks in Appendix \ref{regression-section}), along with ablations to quantify how key architectural components interact with scaling. Across all experiments, our goal is not only to assess HiMAE’s efficiency and transferability, but also to test the \textit{resolution hypothesis}: whether predictive signal concentrates at specific levels of the hierarchical embeddings. Further analysis and results are displayed in full in Appendix Section \ref{addresults}.

\noindent\underline{Model scaling and generative reconstruction:} 

We first study HiMAE’s scaling properties by measuring how reconstruction performance varies as a function of dataset size, number of participants, model capacity, and training compute capacity (batch size). For each axis, we systematically subsample or expand the relevant resource while holding others fixed, enabling us to isolate its contribution to representation quality. Scaling is assessed through mean squared error on masked reconstruction, which provides a direct measure of how model capacity and data availability govern loss reduction. We also squeeze in ablations in this experiment to assess how removing skip connections, and removing the hierarchal design affect scaling.

To complement this aggregate view, we also evaluate generative performance under three increasingly challenging reconstruction regimes defined in the LSM papers \citep{narayanswamyscaling, xu2025lsm}: (i) random imputation, where patches are masked at random uniformly; (ii) temporal interpolation, where contiguous spans are removed to simulate sensor dropout; and (iii) temporal extrapolation, where future spans are occluded and predictions must rely solely on past context.  We compute the mean squared error (MSE) for these evaluations.  

\noindent\underline{Classification:}

To assess downstream transferability and adaptability, we benchmark HiMAE on 12 binary classification tasks drawn from labeled datasets fully disjoint from our pretraining sources. We organize these into three groups: cardiovascular outcomes, sleep staging, and abnormal laboratory prediction.  Cardiovascular outcomes, provide the most established benchmarks, with well-documented links between PPG and clinical endpoints \citep{shabaan2020survey}. These include hypertension detection, estimating blood pressure (blood pressure regression pushed to Appendix \ref{regression} due to poor performance across all models), and arrhythmia-related events such as Premature Ventricular Contractions (PVCs), typically identified via electrocardiograms (ECGs). Sleep staging is another task we include which is of high interest, given the demand for wearables to track fine-grained sleep states despite the temporal and physiological complexity of the task \citep{imtiaz2021systematic, thapa2024sleepfm, birrer2024evaluating}.
Laboratory predictions, on the other hand, serves as a discovery setting, testing whether PPG contains sufficient biomarker information to separate abnormal from healthy labs—an open question compared to patient-record benchmarks where such signals are more explicit \citep{kolo2024meds, arnrich2024medical, mcdermott2025meds}.
 Together, these  canonical and exploratory tasks form a spectrum that enables a comprehensive evaluation of representation quality across diverse digital health applications. All tasks are described in greater detail in Appendix Section~\ref{datasets}.

\begin{figure*}
    \centering
    \includegraphics[width=\linewidth]{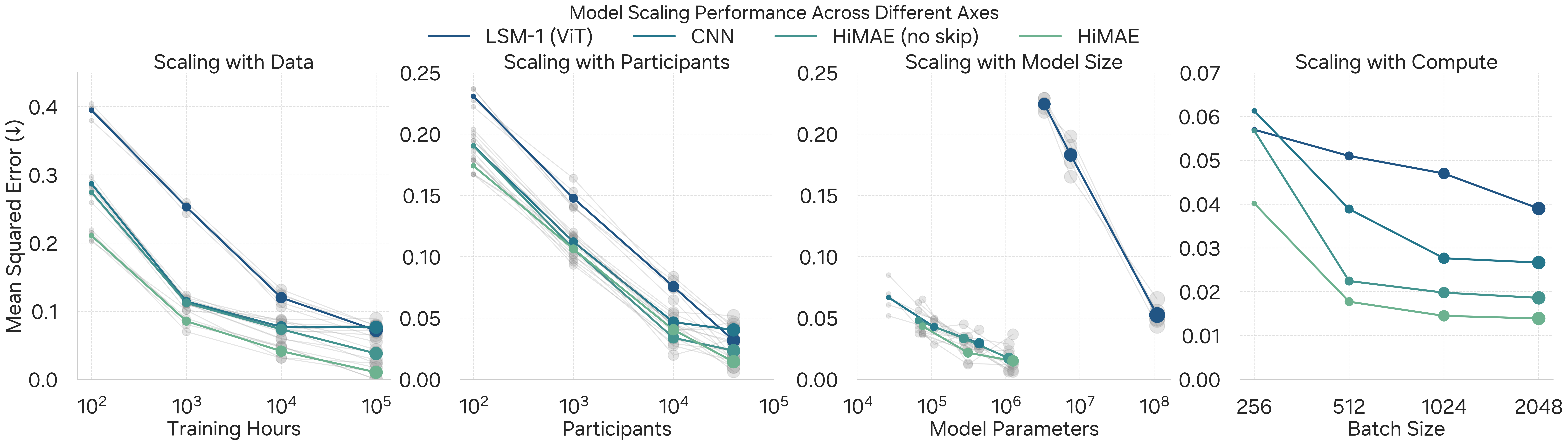}
    \caption{\textbf{HiMAE exhibits superior scaling across axes.} Mean squared error decreases most rapidly for HiMAE as data, participants, model size, and compute scale. Ablations without skip connections confirm that both the hierarchical design and skip pathways are helpful for generative pefromance. Grey lines indicate multiple runs whereas colored lines are average performance. }
    \label{scaling}
\end{figure*}

We compare HiMAE against state-of-the-art SSL methods adapted to the 1D setting for architectural comparability (More details on baselines in Appendix Section \ref{baselines}). Specifically, we include SimCLR~\citep{chen2020simpleframeworkcontrastivelearning}, DINO~\citep{caron2021emergingpropertiesselfsupervisedvision}, Masked Siamese Networks (MSN)~\citep{assran2022masked}, and a hierarchal Swin-Transformer~\citep{liu2021swintransformerhierarchicalvision} as self-supervised baselines, along with the Large Signal Model (LSM)~\citep{narayanswamy2024scaling} and PaPaGei~\citep{pillai2024papagei} as established wearable foundation models. All models are evaluated under standard linear probing, in which the encoder is frozen and a linear classifier is trained on the resulting representations to measure AUROC as the main metric to measure discriminative abilities. For all architectures we use the full sequence embedding across the temporal dimension, without collapsing to a single summary token, to ensure that downstream probes have access to resolution-specific information.
 This setup allows us to test whether pretraining yields representations that are simultaneously transferable across tasks.

\noindent\underline{Resolution Hypothesis:} 

HiMAE produces embeddings at multiple temporal scales, and we probe each scale independently with linear classifiers. This allows us to test whether predictive information is concentrated at fine, intermediate, or coarse resolutions depending on the clinical endpoint. In this way, the classification tasks serve not only as benchmarks for transfer learning, but also as controlled tests of the resolution hypothesis (Receptive field lengths are described in Section \ref{rf-section}). 

\section{Results}

\subsection{Scaling and Generative Benchmark}

\underline{Scaling:} 

We first examine the scaling behavior in Figure~\ref{scaling} of HiMAE relative to baselines across data, participants, model parameters, and compute capacity (batch size). The overall scaling trends follow conventional expectations, error decreases monotonically with additional data, participants, or compute. However, scaling with model parameters reveals a interesting insight. HiMAE achieves substantially lower loss at smaller parameter capacities, while LSMs only begin to close the gap once scaled to orders of magnitude more parameters (we chose LSM parameter count based on their original paper \citep{narayanswamy2024scaling}). This difference reflects an inductive bias. Transformer-based LSMs, which assume global receptive fields, appear to require considerably larger model capacity before capturing the local dynamics of the data (Further Mathematical Intuition is described in Appendix Section ~\ref{extratheory}). In contrast, HiMAE’s hierarchical convolutional structure exploits spatial and temporal locality efficiently, yielding superior performance at modest scales. This observation reinforces the importance of architectural priors in low-capacity regimes.

\noindent\underline{Generative:} 

\begin{figure*}
  \centering
  \vspace{-0.5cm}
  \includegraphics[width=\textwidth]{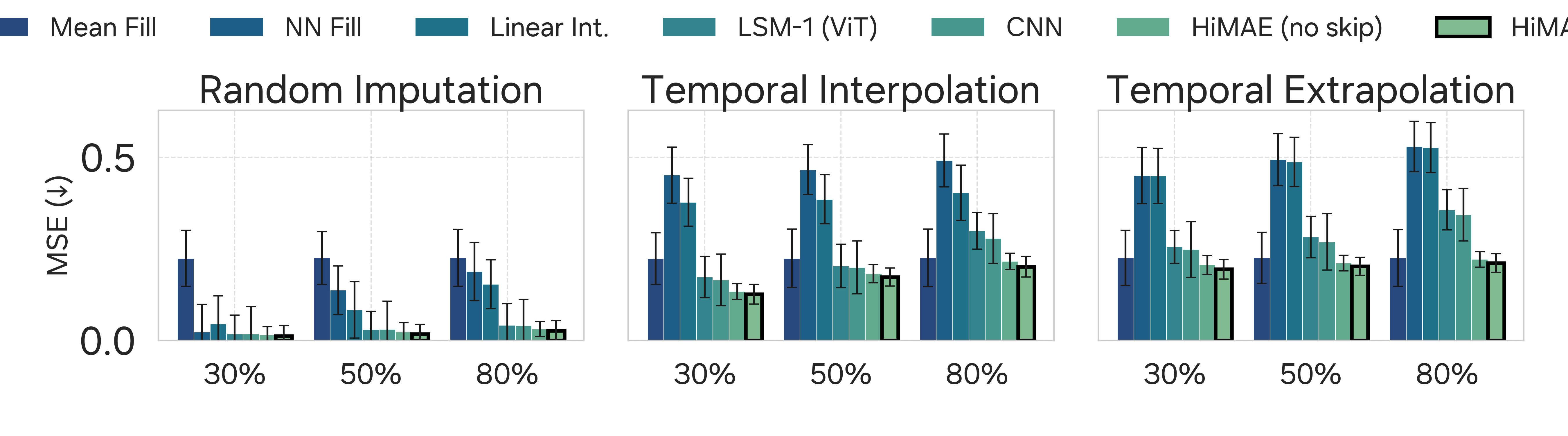}
  \includegraphics[width=\textwidth]{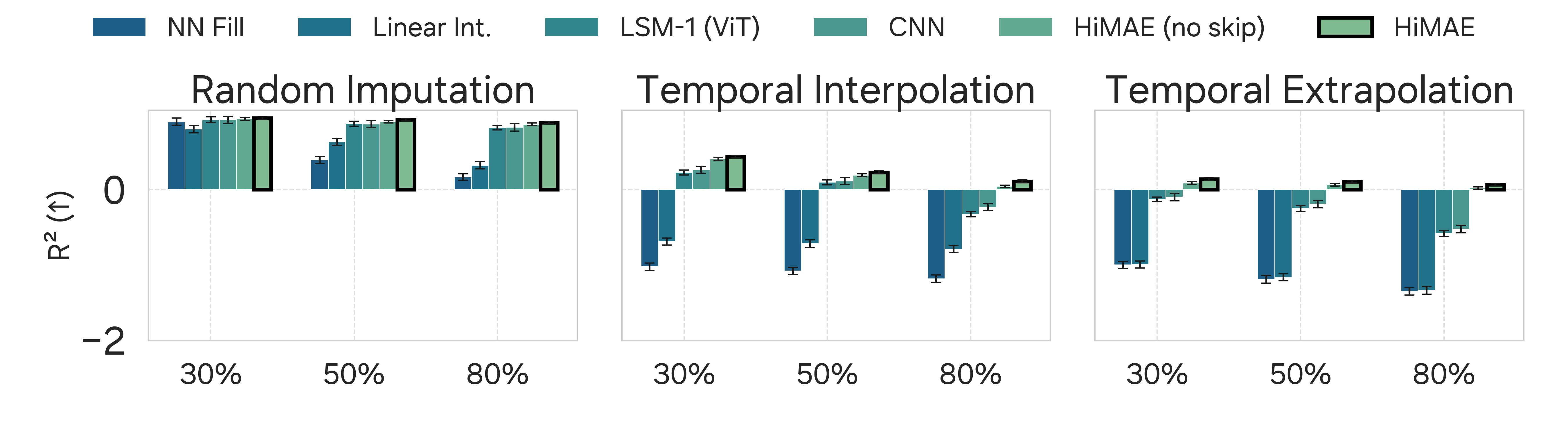}
  \caption{\textbf{Performance on generative benchmarks.} Mean squared error and $R^2$ for random imputation, temporal interpolation, and temporal extrapolation at varying missingness levels. Bold outline indicates best performing model. 
}
  \label{generativebench}
\vspace{-0.5cm}
\end{figure*}

Turning to generative benchmarks, HiMAE consistently outperforms all baselines across random imputation, temporal interpolation, and temporal extrapolation tasks (Figure \ref{generativebench}). In terms of mean squared error, HiMAE achieves the lowest reconstruction error in every setting, including cases with heavy missingness. This advantage persists when evaluated with $R^2$, where the mean-fill baseline serves as the reference. By achieving positive $R^2$ scores even in challenging extrapolation scenarios, HiMAE demonstrates reconstruction ability beyond naive heuristics (e.g., mean fill, nearest neighbor, or linear interpolation). Together, these results establish HiMAE as a strong generative model for missing data problems, with advantages that persist across scaling regimes and input corruption patterns.

\noindent\underline{Ablations:} 

Ablations in Figures \ref{scaling} and \ref{generativebench} further highlights the contributions of hierarchical design and skip connections in HiMAE. Removing either component results in increased error, indicating that both are crucial for effective representation learning. Nevertheless, even without these architectural elements, HiMAE variants remain competitive with larger LSM model, underscoring the robustness of the approach. More importantly, the full model exhibits improved generalization across scaling axes (Appendix Section \ref{scaling-generative}), suggesting that the combination of hierarchy and skip connections facilitates better transfer as data and compute grow.

\subsection{Classification Benchmarking}

\begin{figure*}[t!]
    \centering
    \includegraphics[width=\linewidth]{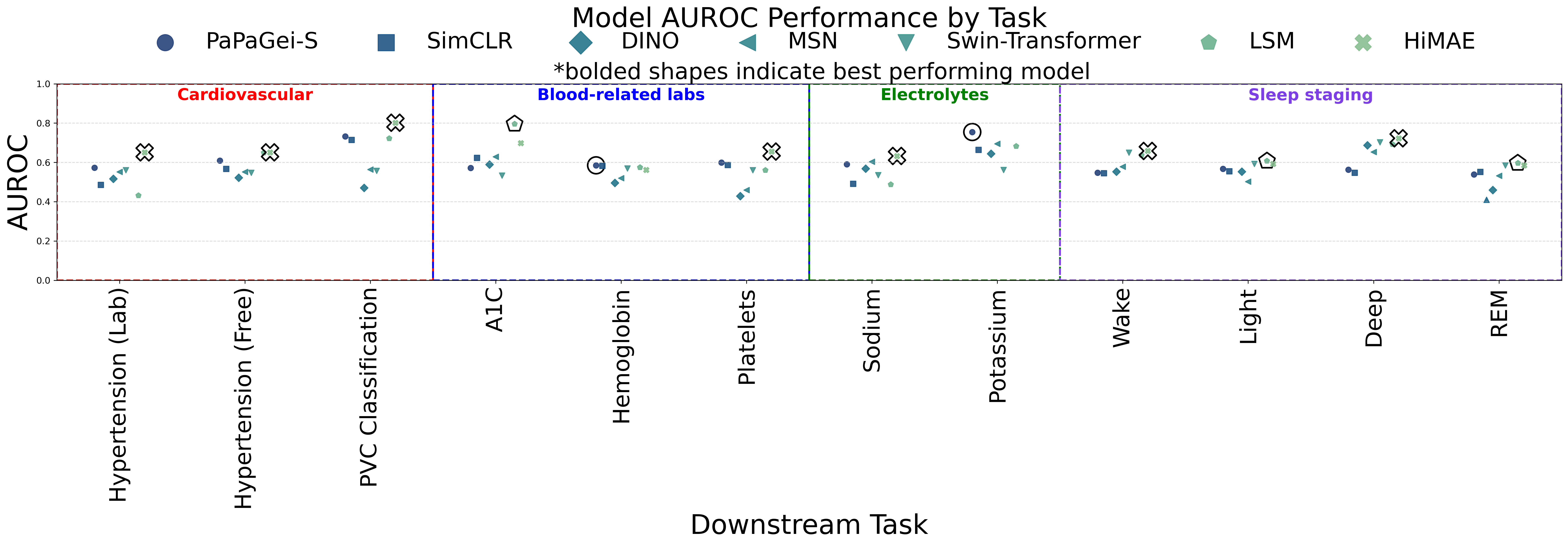}
    \caption{\textbf{AUROC across downstream tasks.} Highlighted shapes indicate best performing model. HiMAE matches or outperforms foundation model baselines with far fewer parameters.}
    \label{classification}
\end{figure*}

In Figure ~\ref{classification}, HiMAE consistently secures the majority of wins, frequently outperforming or matching models that are considerably larger. This is particularly striking given that prior work has typically relied on heavy architectures to reach similar levels of performance, highlighting HiMAE’s ability to capture a broad spectrum of physiological features with a compact design. These outcomes emphasize the model’s robustness when applied to structured, temporally dependent problems that demand sensitivity to subtle variations in wearable signals. 

Taken together, these results position HiMAE as the most consistently strong performer across the benchmark suite. In cases where HiMAE does not place first it is only $\sim$1-2\% behind the winning model. Crucially, this level of performance is achieved with a substantially smaller model than competing approaches, demonstrating a favorable tradeoff between efficiency and predictive power. Rather than excelling only in isolated cases, HiMAE delivers broad, cross-domain competitiveness, suggesting that compact models, when designed with the right inductive biases, can rival or even surpass far larger architectures.

\subsection{Resolution Specific Clinical Interpretability}

\begin{figure*}
    \centering
    \includegraphics[width=\linewidth]{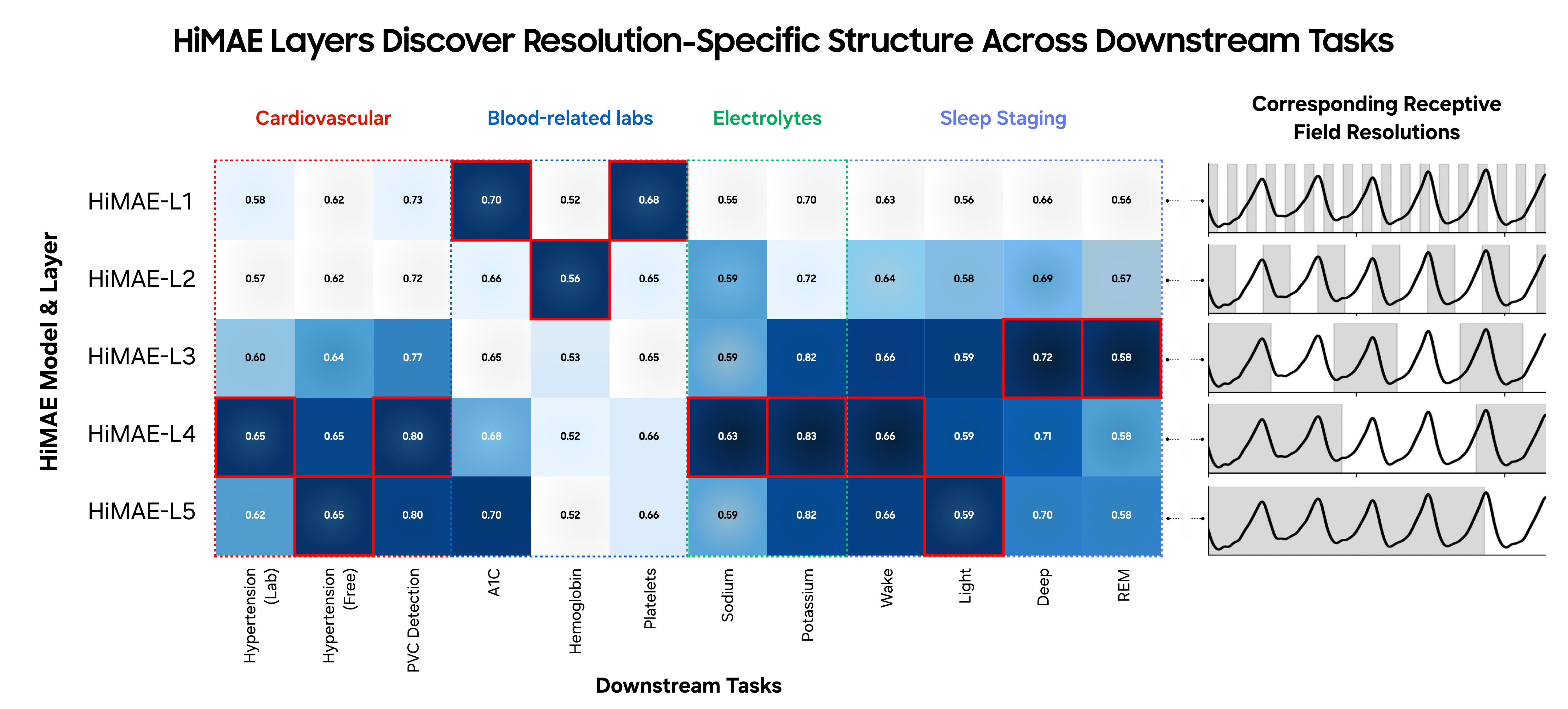}
    \caption{\textbf{HiMAE discovers task-specific structures for downstream tasks.} AUROC across layers shows that tasks rely on distinct temporal scales, highlighting HiMAE as a tool for discovering the most informative resolution in clinical machine learning.
}
    \label{resolution}
\end{figure*}

The resolution hypothesis predicts that different health outcomes depend on distinct temporal granularities. To test this, we analyze performance across HiMAE layers, where each layer corresponds to a progressively coarser resolution. Figure~\ref{resolution} reveals clear resolution-specific structure: individual downstream tasks achieve maximal AUROC at different layers, highlighted by the red boundaries. 

This layer-task alignment underscores two key insights. First, temporal resolution is not a nuisance parameter but an axis of predictive structure: different outcomes are best represented at different scales (we show that collapsing an encoder decoder still has concordant results showing that our hierarchal model is not an artifact in Appendix Section \ref{agreement}). Second, HiMAE naturally exposes this heterogeneity, functioning as a discovery tool for identifying the most informative resolution per task. This complements conventional interpretability methods~\citep{amann2022explain, xu2023interpretability, lee2025clinical} by shifting the focus from \emph{which features} drive predictions to \emph{which resolutions} matter. In doing so, HiMAE operationalizes the resolution hypothesis and provides insights to tasks where the resolution needed is not entirely clear.

\subsection{Case Studies}

\underline{Case Study 1: On-Device Benchmarking}

\begin{wrapfigure}{r}{0.6\textwidth}
  \centering
  \includegraphics[width=0.8\linewidth]{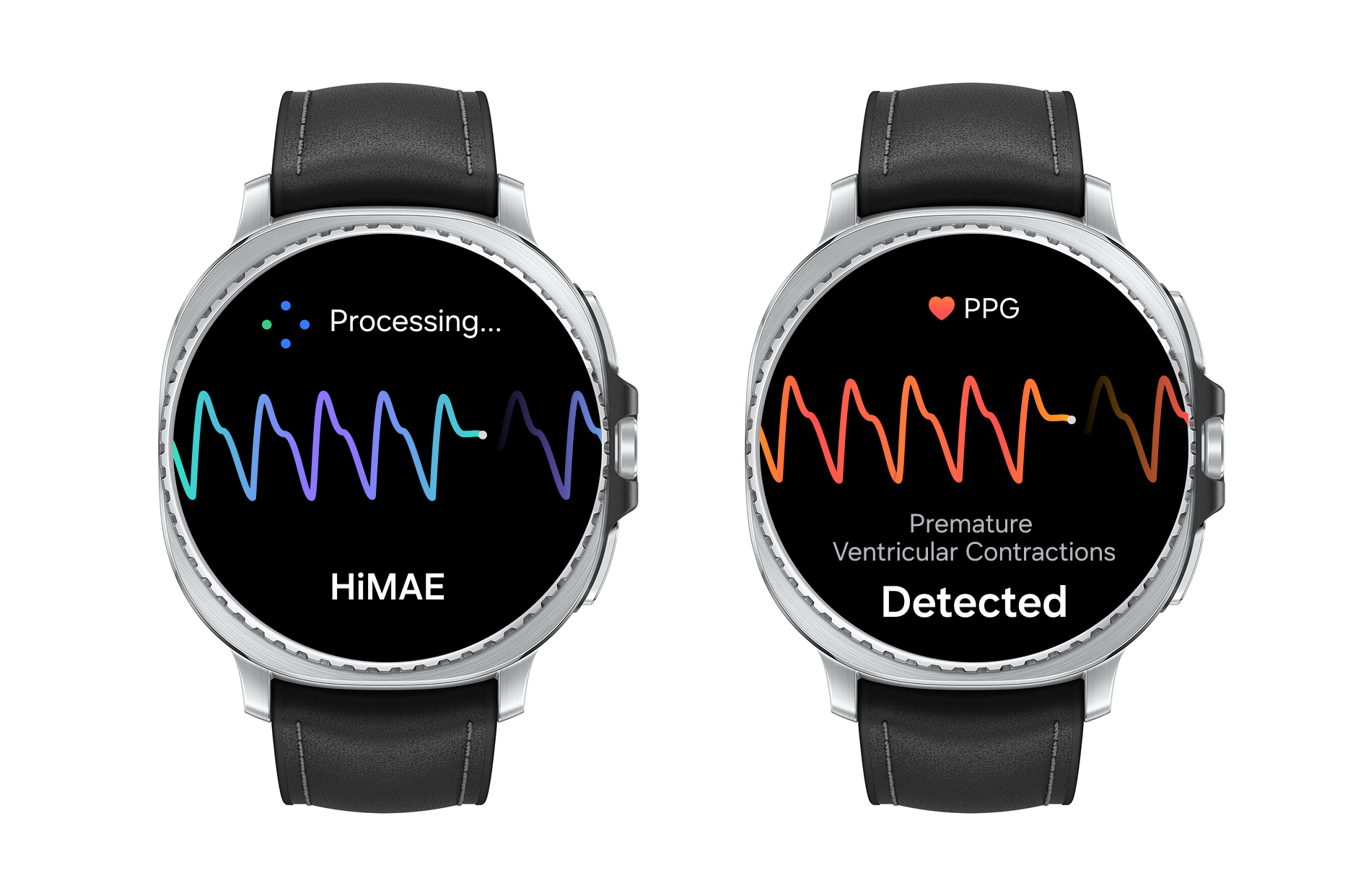}

  \resizebox{0.95\linewidth}{!}{%
    \begin{tabular}{lcccc}
      \toprule
      \textbf{Model} & \textbf{Params ($\downarrow$)} & \textbf{FLOPs ($\downarrow$)} & \textbf{Memory ($\downarrow$)} & \textbf{On-device Lat. ($\downarrow$)} \\
      \midrule
      HiMAE            & \textbf{1.2M }   & \textbf{0.0647} gFLOPs & \textbf{4.8 MB}   & \textbf{0.99 ms} \\
      Efficient-Net B-1 & 7.8M  & 0.70 gFLOPs  & 31.1 MB & 1.42 ms \\
      Swin-Transformer & 110.6M  & 11.89 gFLOPs  & 423.8 MB & 2.95 ms \\
      LSM-Base         & 110.6M  & 15.94 gFLOPs  & 441.3 MB & 3.36 ms \\
      \bottomrule
      \label{watchstats}
    \end{tabular}%
  }
  \caption{\textbf{Model efficiency and on-device inference:} Sample on-device detections on Samsung Watch 8 device. Size, compute cost, memory footprint, and CPU latency (ms per sample, batch size 2048) measured over a 10s sequence at 100Hz.}

  \label{watch}
\end{wrapfigure}

A central novelty of HiMAE is that it is, to our knowledge, the first SSL method compact enough to run entirely \emph{on-watch}, rather than on phone-class hardware. We evaluate on-device PVC detection on smartwatch-class CPUs sampled at 100 Hz (Figure~\ref{watch}). HiMAE is exceptionally lightweight (1.2M parameters, 0.0647 gFLOPs, 4.8 MB) and achieves 0.99 ms latency per sample, equivalent to processing $\approx$1{,}010 samples/s or $\approx$2.8 hours of signal per minute of wall time. By contrast it shows massive performance gains against transformer baselines, Swin-Transformer (110M parameters, 11.9 gFLOPs, 423 MB) and LSM-Base (110M, 15.9 gFLOPs, 441 MB). HiMAE also outperforms optimized models like Efficient-Net B1 \citep{tan2020efficientnetrethinkingmodelscaling} providing context to the latency and compactness of our model. HiMAE is thus $\sim$3–4$\times$ more efficient compared to transformers while fitting fully on-watch (without quantization \citep{jacob2017quantizationtrainingneuralnetworks}), enabling continuous, private inference at the point of signal collection. \textit{This prototype is strictly for research and is not deployed commercially.}
\clearpage
\noindent\underline{Case Study 2: HiMAE is adaptable in few shot settings}

\begin{wrapfigure}{r}{0.6\textwidth}
  \centering
  \includegraphics[width=0.59\textwidth]{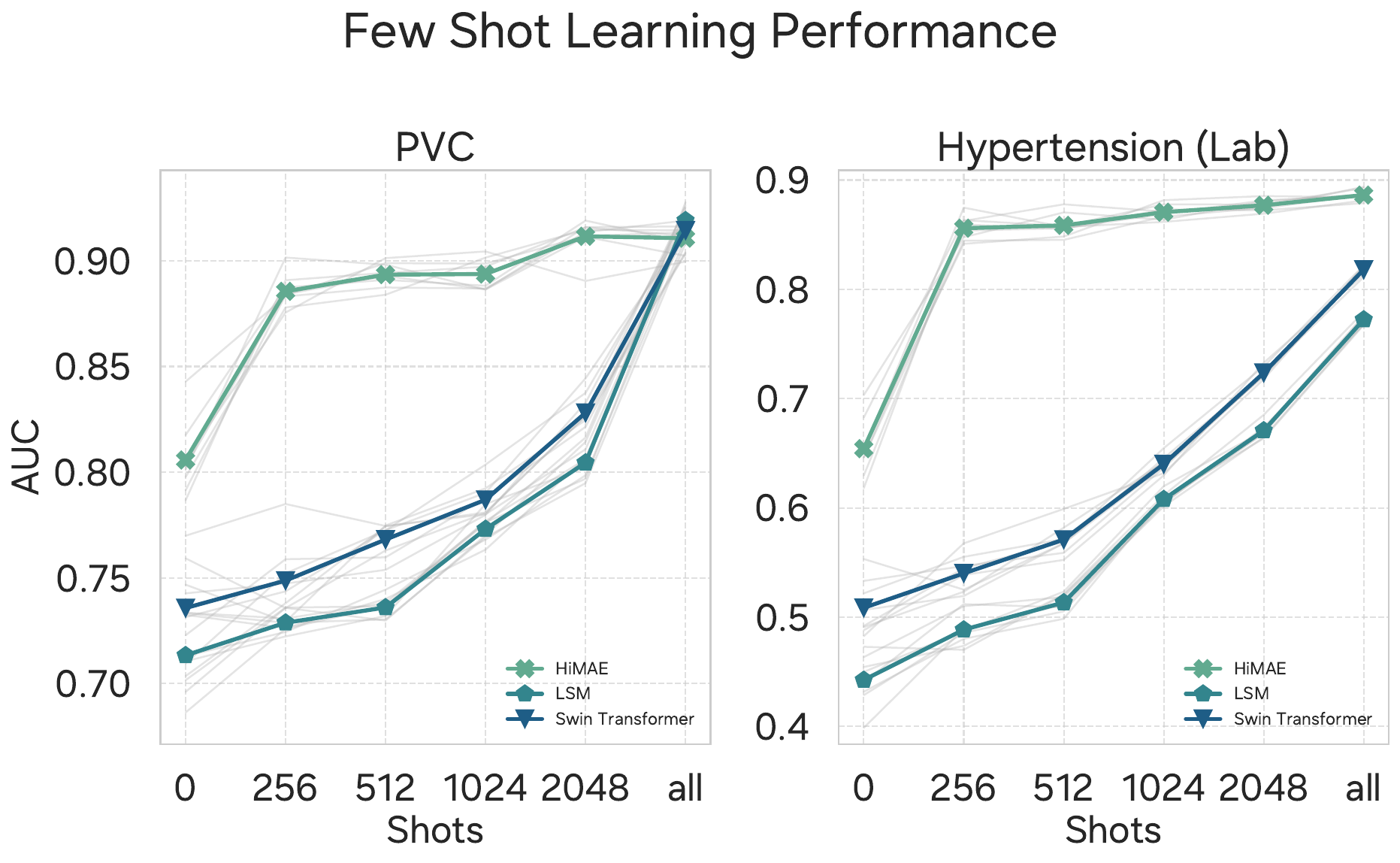}
  \caption{\textbf{Few-shot adaptation.} HiMAE adapts efficiently to new wearable tasks under sparse labels indicated by curve shape over transformer baselines.
}
  \label{fewshot}
\end{wrapfigure}

A central challenge in the wearable domain is that labels are scarce across tasks. Models that can adapt quickly from generic pretraining to specific detection tasks with limited supervision are therefore essential. Figure \ref{fewshot} illustrates this setting: HiMAE provides strong representations that can be adapted to diverse tasks such as PVC detection or hypertension monitoring with only a handful of labeled examples as reflected by the shape of the learning curves on the few-shot learning experiments. By reducing the supervision required to reach high performance, HiMAE enables new tasks to be supported on-device without the prohibitive cost of large curated datasets which help bolster its practical utility. 

\section{Discussion}

\underline{Summary}

HiMAE advances wearable self supervised methods along three dimensions: (i) its flexible architecture is expressly designed for multi-resolution mapping, enabling seamless adaptation across heterogeneous tasks, (ii) by aligning task-dependent resolutions with model representations, it not only optimizes predictive performance but also offers a window into the temporal organization of physiological biomarkers, and (iii) by design of the compactness, it achieves the first demonstration of true \emph{on-watch} inference, running entirely within smartwatch-class constraints while matching or surpassing performance on far larger models. These results position HiMAE as an efficient representation learner but also as a framework for interrogating which temporal resolutions carry signal.

\noindent\underline{Resolution as a structural prior} 

Our findings validate the resolution hypothesis and suggest a shift in how representation learning on wearables should be conceptualized. This reframing implies that representation learning for physiological signals should expose, rather than collapse, scale-specific embeddings. The layer-wise AUROC profiles in Figure ~\ref{resolution} show that predictive performance peaks at different levels of the hierarchy depending on the task, with fine-scale embeddings capturing short-lived physiological events and coarse-scale embeddings capturing slower behavioral phenomena. By revealing this heterogeneity, HiMAE provides empirical evidence that resolution-specific representations are essential for wearable health modeling.

\noindent\underline{From ``on-device'' to ``on-watch.''} 

HiMAE demonstrates that convolutional hierarchies can reduce model size by two orders of magnitude relative to transformer-based LSMs, enabling the first instance of true \emph{on-watch} inference. This moves the deployment frontier from phone-class to watch-class processors, where inference occurs exactly at the point of sensing. Beyond efficiency, this shift has consequences for privacy (data never leave the device) and for clinical viability (continuous real-time monitoring becomes feasible).

\noindent\underline{Limitations and Future Works} 

While we focus on PPG, the principles underlying HiMAE generalize to multimodal settings. Physiological signals are inherently multi-scale across modalities (e.g., ECG beats, accelerometer motion cycles, EEG rhythms), and resolution-aware architectures could expose complementary temporal signatures across them. Another limitation of our work is we don't handle sequences beyond 10 second windows which could unlock another breadth of tasks. Future works also warrants a clinical validation to the discoveries made by HiMAE which could be of significant interest to the health community.

\section*{Acknowledgments}

The Algorithms Team thanks the many contributors who helped bring this project from conception to completion. We are grateful to the researchers and engineers who curated the datasets and designed the methodology, the designers who conceptualized the UI/UX and application interface, and the developers who brought the app to life.

\bibliography{main}

\clearpage
\appendix

\section{Author Contribution}

We attribute proper credit to the following authors for their contributions in this project.

\begin{table}[ht]
\centering
\caption{Overview of author contributions.}
\label{tab:author_contributions}
\renewcommand{\arraystretch}{1.2}
\begin{adjustbox}{max width=\textwidth}
\begin{tabular}{@{}l|ccccccccc@{}}
\toprule
\textbf{Author} & \textbf{Concept} & \textbf{Experiment Design} & \textbf{Coding} & \textbf{Analysis} & \textbf{Writing} & \textbf{Visualization} & \textbf{Project Mgmt.} & \textbf{Discussion} & \textbf{Resources} \\ 
\midrule
Simon A. Lee              &   \checkmark  & \checkmark & \checkmark & \checkmark & \checkmark & \checkmark & \checkmark &          \checkmark   &  \\
Cyrus Tanade          &     &             &     \checkmark        &   \checkmark          &       \checkmark      &             &         \checkmark    & \checkmark & \checkmark \\
Hao Zhou         &     &             &     \checkmark        &   \checkmark          &             & \checkmark             &            & \checkmark &  \\
Juhyeon Lee       &     &             &             &   \checkmark          &       \checkmark       &             &            & \checkmark & \checkmark \\
Megha Thurkal &     &             &             &   \checkmark          &             &             &            & \checkmark &  \checkmark \\
Minji Han &     &             &             &             &             &  \checkmark           &            & \checkmark &  \checkmark \\
Rachel Choi &     &             &             &             &             &  \checkmark           &            & \checkmark &  \checkmark \\
Md Sazzad Hissain Khan &     &             &          \checkmark   &   \checkmark          &             &             &            & \checkmark &  \checkmark \\
Baiying Liu     &             &             &             &    \checkmark          &             &  &             &        \checkmark     &             \\
Keum San Chun               &             &             &             &             &             &  &             &        \checkmark     &     \checkmark        \\
Migyeok Gwak               &             &             &             &             &             &  &             &        \checkmark     &     \checkmark        \\
Mehrab Bin Morshed      &             &             &             &             &             &  &             &        \checkmark     &            \\
Viswam Nathan       &             &             &             &             &             &  &             &        \checkmark     &             \\
Mahbubur Rahman      &             &             &             &             &             &  &             &        \checkmark     &     \checkmark        \\
Li Zhu       &             &             &             &             &             &  &             &        \checkmark     &           \\
Subramaniam Venkatraman      &             &            &             &             &             &  &      \checkmark       &        \checkmark     &     \checkmark        \\
Sharanya Desai      &             &      \checkmark       &             &             &   \checkmark          &  &      \checkmark       &        \checkmark     &     \checkmark        \\
\bottomrule
\end{tabular}
\end{adjustbox}
\end{table}
\section{Frequently Asked Questions}

\noindent \underline{What are the main conclusions from this work?}  

Our contributions are twofold and interdependent: first, we introduce a compact convolutional model whose inductive bias drives both efficiency and robustness; second, we show that this compactness enables the first smartwatch-level deployment of PPG inference without reliance on phone processors. Each contribution reinforces the other—compact inductive design is what makes on-device deployment feasible, and on-device feasibility highlights the practical impact of our design. We demonstrate that convolutional architectures indeed benefit from inductive biases that remain advantageous for PPG signals. On our pre-training data, our model consistently outperforms alternative baselines. Scaling experiments across model sizes further reveal that while brute-force scaling of generic architectures is possible, it is less effective: our model achieves stronger performance and scales more gracefully owing to better initialization and inductive structure.

\noindent\underline{Is your pre-training dataset large enough?}  

Our pre-training corpus was collected internally and is of comparable scale to recent public benchmarks such as PaPaGei and Apple’s datasets. In terms of magnitude, we position our dataset as \(\text{PaPaGei \citep{pillai2025papageiopenfoundationmodels}} < \text{Ours} < \text{Apple \citep{abbaspourazad2023large}} < \text{Google \citep{narayanswamy2024scaling}}\). Thus, while not the largest available, our dataset size is sufficiently large to validate the approach and lies within the range of accepted practice for self supervised learning wearable models.

\noindent\underline{Why do you model at 10-second windows?}  

We deliberately adopt 10s windows sampled at 100Hz to balance physiological coverage with on-device feasibility. Many clinically actionable events, such as arrhythmic beats or premature ventricular contractions, unfold on the order of seconds and require rapid detection to enable continuous monitoring and real-time feedback. Shorter windows would impair the model’s ability to capture meaningful temporal context, while much longer windows would hinder low-latency inference on watch-class hardware. By constraining the receptive field to 10s, HiMAE preserves second-level resolution while remaining efficient enough to process signals continuously under the hardware limits of edge devices. Additionally, 10-second window are a standard protocol that are adopted in the clinical setting where ECG for example is collected and interpreted at 10 second segments \citep{shuai201610}.

\noindent\underline{What are the advantages of smaller models?}  

From a research perspective, smaller models foster inclusivity by reducing reliance on brute-force scaling of transformer-based architectures that only industry-scale labs can realistically afford. From a deployment standpoint, compact models enable on-device inference on constrained hardware such as wearables. This dual benefit (lower research barriers and wider deployment potential) underscores the importance of investigating architectures that remain competitive at modest scale.

\noindent \underline{How large is too large to deploy on a smart watch?}  

In principle, models up to approximately 50MB can be stored and executed on modern smart watches or larger models can be quantized \citep{jacob2017quantizationtrainingneuralnetworks}. In practice, however, latency and energy considerations suggest that models exceeding roughly 10MB may already hinder real-time inference and limit commercial viability. Additionally quantization does not do due dilligence to the original model and some level of the model's performance is lost. While smartphones relax these constraints, our contribution highlights that the proposed model remains sufficiently compact to fit within the computational and storage budgets of wearable devices such as watches, thereby supporting direct on-device deployment.

\noindent\underline{Can PPG predict abnormal laboratory results?}

While exploratory and not clinically actionable, these findings highlight the role of AI not only as a predictive tool but also as a means of discovery in biomedical science. We investigate whether photoplethysmography (PPG) signals encode latent biomarkers that distinguish ``normal'' from ``abnormal'' lab values. Using lightweight classifiers on frozen embeddings with strict temporal alignment, we probe whether learned PPG representations capture physiological signatures correlated with out-of-range labs. Preliminary evidence suggests discriminative signal above chance, pointing to the possibility that AI can surface hidden biomarkers and reveal new aspects of human physiology.

\clearpage
\section{Ethics Considerations}
\label{ethics}

\subsection{Data Privacy and Consent}
Wearable signals capture sensitive physiological and behavioral information \citep{erturk2025beyond}. Our study relies on both publicly available and proprietary (company-owned) datasets that have been carefully vetted. These datasets include transparent disclosure of data usage, explicit opt-in mechanisms, and the option for participants to withdraw \citep{perez2021digital}. Across the seven datasets used in this study, we obtained written consent—via paper or digital waivers—that clearly informed participants that their data may be used for commercial research purposes.

\subsection{Bias and Representativeness}
Physiological signals vary across age, gender, ethnicity, health status, and socioeconomic context, yet most existing datasets underrepresent key populations \citep{fitzgerald2017implicit, mccradden2020patient, chen2021ethical}. Such underrepresentation risks embedding biases into foundation models, leading to inequitable performance in downstream applications. Mitigation requires deliberate corpus curation, bias auditing, and systematic evaluation across diverse cohorts. In this study, we sought to mitigate bias by incorporating a pre-training corpus drawn from a wide range of wearable devices, collected across multiple regions of the world and over many years. However, patient-specific demographic information is not available. We do note that our data was collected across 4 countries including, USA, Brazil, Bangladesh, and South Korea.

\subsection{Clinical Implications}
Wearable foundation models are not substitutes for medical judgment. Their predictions require regulatory approval and clinical validation before integration into healthcare practice. Without safeguards, model misinterpretation could lead to misdiagnosis or inappropriate treatment. Development should involve clinical collaborators, real-world evaluations, and explicit positioning of models as decision-support rather than diagnostic systems. In our group, ongoing collaborations aim to evaluate where our foundation model performs well and how it may assist in forming clinical insights. We emphasize that no definitive clinical conclusions should be drawn from this work.

\subsection{Environmental Impact}
Training generative models entails substantial computational and environmental costs \citep{Ligozat_2022, stochasticparrot, Bouza_2023}. To minimize our footprint, we limited redundant runs, and reused checkpoints to avoid unnecessary GPU usage. All experiments were conducted on datacenter GPUs with efficient cooling systems and renewable energy credits to reduce carbon intensity. We emphasize that transparent reporting of compute usage and bounding resource allocation are necessary steps toward sustainable machine learning research.

\clearpage
\section{Reproducibility Statement}

\begin{table*}[h!]
\centering
\caption{HiMAE architecture components.}
\label{tabhimaecomponents}

\begin{minipage}{0.45\linewidth}
\centering
\caption*{Encoder--Decoder}
\begin{adjustbox}{max width=\linewidth}
\begin{tabular}{ll}
\toprule
\textbf{Layer} & \textbf{Output Shape} \\
\midrule
Input & [B, 1, $T$] \\
EncoderConvBlock(1$\to$16) & [B, 16, $T$/2] \\
EncoderConvBlock(16$\to$32) & [B, 32, $T$/4] \\
EncoderConvBlock(32$\to$64) & [B, 64, $T$/8] \\
EncoderConvBlock(64$\to$128) & [B, 128, $T$/16] \\
EncoderConvBlock(128$\to$256) & [B, 256, $T$/32] \\
DecoderSkipBlock(256$\to$128) & [B, 128, $T$/16] \\
DecoderSkipBlock(128$\to$64) & [B, 64, $T$/8] \\
DecoderSkipBlock(64$\to$32) & [B, 32, $T$/4] \\
DecoderSkipBlock(32$\to$16) & [B, 16, $T$/2] \\
Final Deconv (16$\to$1) & [B, 1, $T$] \\
Tanh & [B, 1, $T$] \\
\bottomrule
\end{tabular}
\end{adjustbox}
\end{minipage}%
\hfill
\begin{minipage}{0.25\linewidth}
\centering
\caption*{EncoderConvBlock}
\begin{adjustbox}{max width=\linewidth}
\begin{tabular}{l}
\toprule
\textbf{Layer} \\
\midrule
Conv1d ($k=5$, s=2, p=2) \\
BatchNorm \\
GELU \\
Conv1d ($k=5$, s=1, p=2) \\
BatchNorm \\
Conv1d ($k=1$, s=2) + BN \\
GELU \\
\bottomrule
\end{tabular}
\end{adjustbox}
\end{minipage}%
\hfill
\begin{minipage}{0.25\linewidth}
\centering
\caption*{DecoderSkipBlock}
\begin{adjustbox}{max width=\linewidth}
\begin{tabular}{l}
\toprule
\textbf{Layer} \\
\midrule
ConvTranspose1d ($k=5$, s=2, p=2, op=1) \\
Concat skip connection \\
Conv1d ($k=5$, s=1, p=2) \\
BatchNorm \\
GELU \\
Conv1d ($k=5$, s=1, p=2) \\
BatchNorm \\
GELU \\
\bottomrule
\end{tabular}
\end{adjustbox}
\end{minipage}
\label{tabhimaecomponents}
\end{table*}

Due to restrictions around data licensing and industry policies, we are unable to release the full source code associated with HiMAE. To mitigate this limitation, we provide complete details of the model architecture, layer configurations, and hyperparameters in Table~\ref{tabhimaecomponents}. This includes all encoder, decoder, and skip connection blocks, along with kernel sizes, strides, padding, activation functions, and normalization layers. Together, these descriptions are sufficient to re-implement the model faithfully in any modern deep learning framework \citep{paszke2019pytorchimperativestylehighperformance, 10.5555/3026877.3026899, jax2018github, mlx2023}. In addition, we report all training settings (e.g., optimizer, learning rate schedule, and batch size) in the Appendix Section \ref{baselines} to further support reproducibility. Our goal is to ensure that, while the exact implementation cannot be shared, independent researchers can replicate the methodology and validate the findings presented in this work.

\subsection{Temporal Resolution and Receptive Field}
\label{rf-section}

\begin{table}[h]
\centering
\caption{Temporal resolution and cumulative receptive field through the encoder. $T$ denotes the input length in samples. $R_\ell$ is the receptive field after layer $\ell$ and $J_\ell$ the effective input stride (“jump”).}
\label{recept}
\begin{adjustbox}{max width=\linewidth}
\begin{tabular}{lcccc}
\toprule
Layer & Kernel $k$ & Stride $s$ & Output length & $R_\ell$ / $J_\ell$ \\
\midrule
Enc1-conv1 & 5 & 2 & $T/2$  & $5$   / $2$  \\
Enc1-conv2 & 5 & 1 & $T/2$  & $13$  / $2$  \\
Enc2-conv1 & 5 & 2 & $T/4$  & $21$  / $4$  \\
Enc2-conv2 & 5 & 1 & $T/4$  & $37$  / $4$  \\
Enc3-conv1 & 5 & 2 & $T/8$  & $53$  / $8$  \\
Enc3-conv2 & 5 & 1 & $T/8$  & $85$  / $8$  \\
Enc4-conv1 & 5 & 2 & $T/16$ & $117$ / $16$ \\
Enc4-conv2 & 5 & 1 & $T/16$ & $181$ / $16$ \\
Enc5-conv1 & 5 & 2 & $T/32$ & $245$ / $32$ \\
Enc5-conv2 & 5 & 1 & $T/32$ & $373$ / $32$ \\
\bottomrule
\end{tabular}
\end{adjustbox}
\end{table}


\clearpage
\section{Datasets}

\label{datasets}

\subsection{Aquistion and approval}

All data analyzed in this study were collected under informed consent, with participants explicitly agreeing for their wearable-derived signals to be used in health-related research. The consent language stated that data could be used for developing new health features and algorithms and for inclusion in scientific publications. In particular, participants were informed that health and wellness data such as steps, heart rate, sleep, and photoplethysmography (PPG) signals could contribute to findings aimed at advancing general knowledge of health and science. No data used in this study included personally identifying information such as names or email addresses. We attach a portion of the protocols defined in our user data agreements below:

\textit{The use of these de-identified data for data usage was reviewed and classified as exempt. In addition, because the supporting records constitute case histories and document exposure to devices, we complied with the recordkeeping requirements in 21~CFR~\S~812.140(a)(3), including obtaining written digital consent and dated information. Participants could withdraw at any time; such withdrawals were documented in the case history, and data collected up to the point of withdrawal were retained and used for the investigation in accordance with the consent and applicable regulations.}

For downstream evaluations, we relied on a combination of institutional review board (IRB)-approved datasets and publicly available resources. For instance, the PVC detection task used paired PPG and ECG recordings to derive annotations of premature ventricular contractions, with ECG-based labels verified both algorithmically and manually. The hypertension classification tasks were drawn from the My Heart  Lab Study collected in a lab Setting (ID NCT04314947) and My BP Lab (Clinical Trials ID 19-27169) studies collected in a free-world settting, both of which collected wrist-based PPG alongside reference blood pressure measurements under IRB-approved protocols. Sleep staging was evaluated using the DREAMT dataset, which combines PPG with gold-standard polysomnography annotations in individuals with and without diagnosed sleep disorders. Finally, a range of abnormal lab test prediction tasks were derived from the Tulane University dataset (ID 20242033), linking PPG from Samsung devices with clinical laboratory values for biomarkers (More details in Appendix Section \ref{datasets}). 

Across all studies, participants consented to data collection through mobile platforms that supported eligibility screening and enrollment, provided full informed consent, and enabled seamless integration of Samsung devices for continuous signal acquisition. Where appropriate, participants also reported medical histories or completed questionnaires through these platforms. All data were de-identified and stored in accordance with the approved study protocols, ensuring compliance with ethical and regulatory standards.

This layered consent and governance framework ensures that the data underpinning our pretraining and evaluation tasks are both ethically sourced and scientifically robust, supporting the broader goal of advancing health monitoring through consumer wearables such as the \textcolor{red}{REDACTED} Watch.

\subsection{Pre-training and Generative Datasets}

\subsubsection{Device Distribution}

\begin{figure}[htbp]
    \centering
    \includegraphics[width=0.6\linewidth]{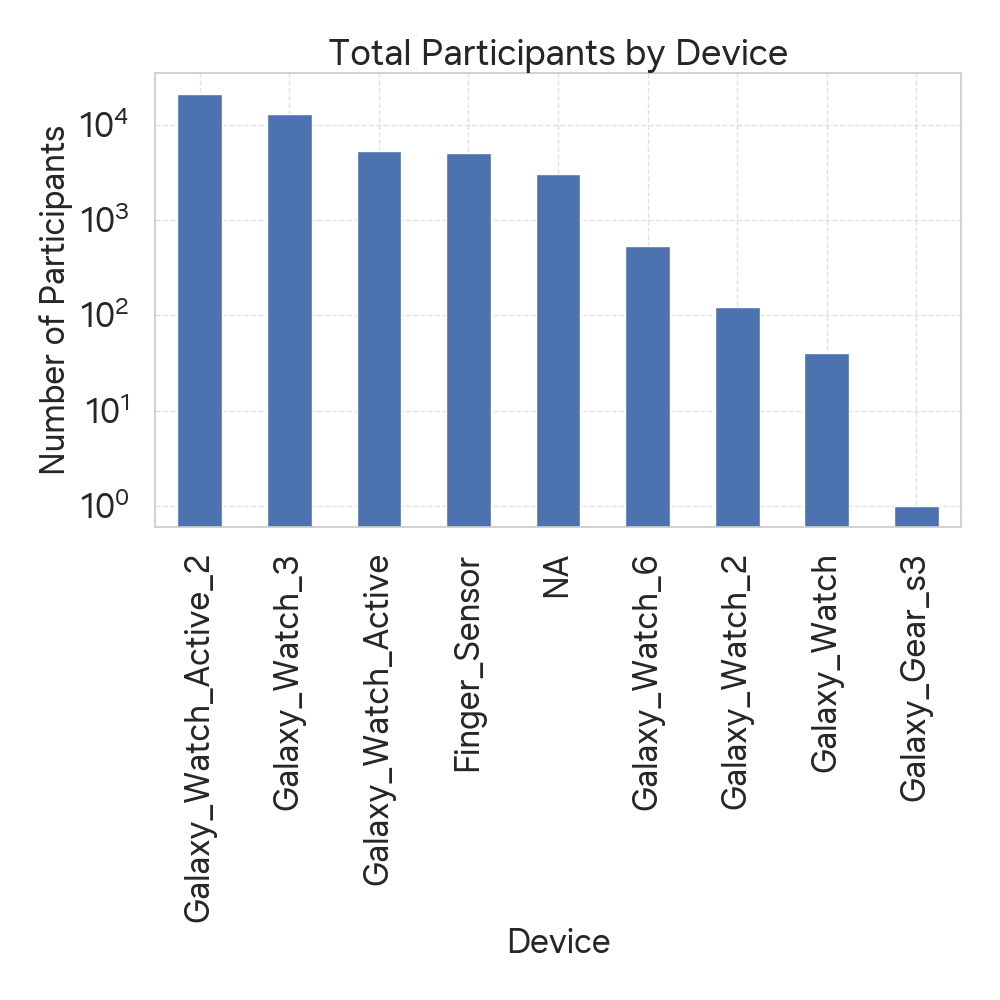}
    \caption{\textbf{Total Participants by Device.} The figure displays a bar chart illustrating the distribution of participants across different wearable devices used in the study. The y-axis is on a logarithmic scale to better show the wide range in the number of participants}
    \label{typeofwatch}
\end{figure}

The distribution of participants and data availability highlights both the diversity of collection devices and the heterogeneity of study contributions (Figure \ref{typeofwatch}). At the device level, participation is primarily sourced from Galaxy  Watch Active 2, Galaxy Watch 3, Galaxy Watch Active, each contributing a lot of  participants, while older models such as the Galaxy Gear S3 
are represented by fewer users. This heterogeneity in devices provide us with a realistic and diverse set of raw wearable signals that can help us build generalizable foundation models. The presence of entries labeled as “NA” further reflects the mixture of collection devices and the occasional incompleteness of metadata. \textit{We note that the devices used in our study are provided by two distributors limiting its generalizability and causing potential biases due to not having access to other consumer wearable devices.}

\subsubsection{Participant Counts}

\begin{figure}[htbp]
    \centering
    \includegraphics[width=0.6\linewidth]{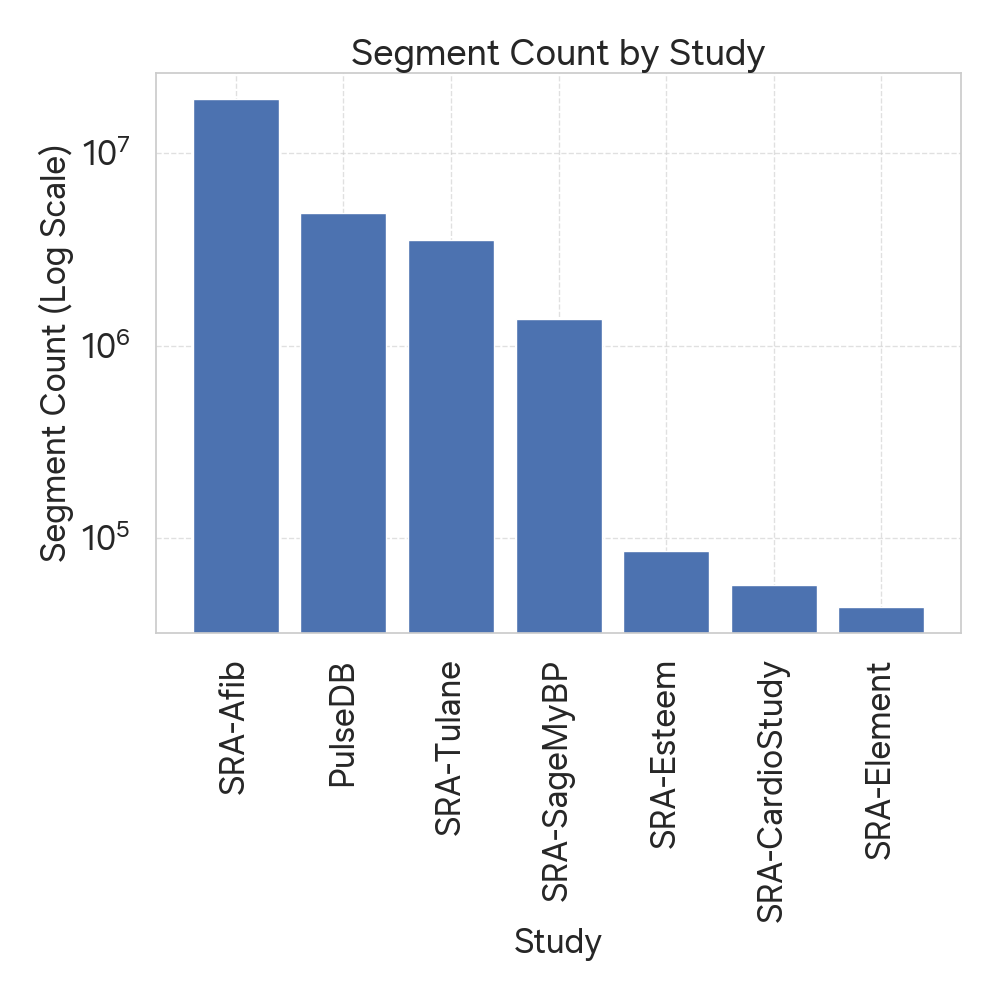}
    \caption{\textbf{Segment Count by Study.} This bar chart shows the number of data segments collected for each study, with the y-axis on a logarithmic scale to account for the large differences in segment counts.}
    \label{study}
\end{figure}

In terms of study based segmentation, the dataset contains a handful of large-scale cohort studies, leading to diverse representation (Figure \ref{study}). Efforts were made to ensure representation across studies of varying sizes. This underscores the necessity of leveraging the vast scale of high-volume cohorts while simultaneously preserving the heterogeneity introduced by smaller studies, since both dimensions are essential for building foundation models that truly capture the variability and complexity of one-dimensional PPG signal modeling. Our data was collected across 4 countries (USA, South Korea, Brazil, Bangladesh) though most people specific demographic information is missing.


\subsubsection{Pre-processing pipeline}

We operate on fixed-length windows (10 s) of raw PPG sampled at device-specific rates $f_s$. Each window is standardized via per-window z-scoring, $\tilde{x}_t=(x_t-\mu)/\sigma$, to remove level and scale effects that confound morphology-based quality metrics. To suppress gross amplitude artifacts (e.g., motion bursts), we compute the skewness of $|\tilde{x}|$, denoted $\gamma=\mathrm{skew}(|\tilde{x}|)$. Windows with heavy-tailed amplitude distributions ($\gamma>2$) undergo an iterative trimming procedure that discards high-percentile excursions and recomputes $\gamma$ until the distribution regularizes or a conservative floor is reached. This stage intentionally trades recall for precision: if trimming fails to regularize the distribution, the window is rejected.

For windows that pass amplitude checks, we impose a regularity prior using the sample autocorrelation $r[k]=\sum_t \tilde{x}_t \tilde{x}_{t+k}$. We locate zero-crossings of $r[k]$ near the origin and compute the dispersion of consecutive intervals, $\sigma_{\mathrm{zc}} = \mathrm{std}(\Delta k)/f_s$. Physiologically plausible pulsatile signals exhibit near-periodic structure; we therefore require a small timing dispersion to proceed. This criterion rejects segments whose periodicity is unstable, a signature of motion or sensor decoupling, and eliminates short or degenerate traces by enforcing a minimum number of intervals.

Surviving windows are band-limited with a low-order Butterworth filter to the cardiac band $[0.1, 2]$ Hz, which removes drift and high-frequency noise without distorting pulse morphology. We then quantify morphology via template matching against a canonical PPG waveform. Let $q_t\in[0,1]$ denote the per-sample similarity score. We define a stringent acceptance mask $m_t=\mathbf{1}\{q_t>\tau\}$ with $\tau\in\{0.90,0.95\}$ depending on whether the amplitude distribution was already regular ($\gamma\le 2$). Two complementary statistics summarize quality: a “coverage” term $p=\frac{1}{T}\sum_t m_t$, measuring the fraction of the window that is confidently PPG-like, and an “agreement” term $a=\frac{1}{\max(1,\sum_t m_t)}\sum_t q_t m_t$, measuring how well accepted samples match the template. To penalize windows that have high agreement on vanishing coverage (or vice versa), we aggregate with the harmonic mean $H(a,p)=\frac{2ap}{a+p}$, yielding a continuous signal-quality index. A small additive term encodes whether the amplitude distribution was regular at entry, prioritizing windows that never required trimming. Windows that fail any upstream gate (amplitude regularization, periodicity stability, or template evaluation) are assigned null quality and excluded from downstream training.

At corpus scale, we apply this scoring in parallel and retain only windows with high composite quality. The resulting pretraining set emphasizes clean, consistent, periodic, and band-pass filtered signals harmonizing across devices and sampling rates, reducing the prevalence of motion artifacts and non-physiologic segments without relying on patient-level demographics or labels.

\subsection{Downstream Evaluation Data}

We evaluate HiMAE across diverse downstream tasks to assess the generality of wearable PPG representations. Rather than assuming a fixed mapping between PPG and outcomes, we exploit HiMAE’s ability to learn hierarchical temporal features and adaptively resolve signal segments at scales most informative for prediction. This design allows us to probe the representational value of optical physiological signals across clinically and behaviorally relevant applications.

\subsubsection{PVC Detection}

\begin{table}[h!]
\centering
\caption{Stratified 80/20 Train/Test splits for PVC tasks (with per-task totals).}
\label{pvcdata}
\begin{adjustbox}{max width=\linewidth}
\begin{tabular}{llrrr}
\toprule
Task & Split & Negative & Positive & Total \\
\midrule
\multirow{3}{*}{PVC Detection} & train & 369987 (91.8\%) & 32832 (8.2\%) & 402819 \\
 & test & 69880 (89.7\%) & 8019 (10.3\%) & 77899 \\
 & totals & 439767 (91.4\%) & 40950 (8.6\%)& 480717 \\
\addlinespace
\bottomrule
\end{tabular}
\end{adjustbox}
\end{table}

Premature Ventricular Contractions (PVCs) (Number Breakdowns in Table \ref{pvcdata}) are abnormal beats arising in the ventricles \citep{cha2012premature, kaya2015classification}. We use paired PPG–ECG data, with ECG annotations generated using BeatLogic \citep{teplitzky2020deep} and manually verified. PPG inputs are 10s non-overlapping wrist segments, pre-processed with a Savitzky–Golay filter \citep{luo2005savitzky}, a $0.5$–$4.0$ Hz bandpass, normalization to $[-1,1]$, and exclusion of segments with motion artifacts or disruptions $>1$ s. This task evaluates whether ubiquitous PPG can approximate arrhythmia detection typically restricted to ECG.

\subsubsection{Hypertension Classification}

\begin{table}[h!]
\centering
\caption{Stratified 80/20 Train/Test splits for Hypertension tasks collected in a laboratory setting.}
\label{element}
\begin{adjustbox}{max width=\linewidth}
\begin{tabular}{llrrr}
\toprule
Task & Split & Negative & Positive & Total \\
\midrule
\multirow{3}{*}{Hypertension Classification (Lab)} & train & 2964 (86.7\%) & 454 (13.3\%) &  3418 \\
 & test & 631 (76.7\%) & 192 (23.3\%) & 823 \\
 & totals & 3595 (84.8\%)  & 646 (15.2\%) & 4241 \\
\addlinespace
\bottomrule
\end{tabular}
\end{adjustbox}
\end{table}

\begin{table}[h!]
\centering
\caption{Stratified 80/20 Train/Test splits for Hypertension tasks collected in a free-world setting.}
\label{sagemybp}
\begin{adjustbox}{max width=\linewidth}
\begin{tabular}{llrrr}
\toprule
Task & Split & Negative & Positive & Total \\
\midrule
\multirow{3}{*}{Hypertension Classification (Free World)} & train & 3959 (58.5\%) & 2812 (41.5\%) & 6771 \\
 & test & 1042 (58.8\%) & 731 (41.2\%) & 1773 \\
 & totals & 5001 (58.5\%) & 3543 (41.5\%) & 8544 \\
\addlinespace
\bottomrule
\end{tabular}
\end{adjustbox}
\end{table}

Hypertension classification (Number Breakdowns in Tables \ref{element}, \ref{sagemybp}) relies on cuff-based references \citep{simonneau2004clinical, giles2005expanding, giles2009definition, simonneau2009updated, simonneau2013updated, simonneau2019haemodynamic}. Subjects within $\pm 8$ mmHg of the diagnostic cutoff are excluded to reduce label noise, with remaining individuals labeled hypertensive or normotensive. Each 10s PPG segment undergoes Savitzky–Golay smoothing, $0.5$–$4.0$ Hz bandpass filtering, normalization to $[-1,1]$, and artifact removal. Unlike PVC detection, which is event-based, this task leverages PPG morphology and temporal dynamics to reflect vascular state. These evaluations contain both hypertension data collected in a naturalistic free world environment and within a controlled lab environment for both the hypertensive and blood pressure regression tasks.

\subsubsection{Sleep Staging}

\begin{table}[h!]
\centering
\caption{Stratified 80/20 Train/Test splits for DREAMT Dataset Sleep Staging.}
\label{sleep}
\begin{adjustbox}{max width=\linewidth}
\begin{tabular}{llrrrrr}
\toprule
Task & Split & Wake & Light & Deep & REM & Total \\
\midrule
\multirow{3}{*}{Sleep Staging (4-class)} & train & 44829 (23.9\%) & 115932 (61.8\%) & 6696 (3.6\%) & 20214 (10.8\%) & 187671 \\
 & test & 11298 (23.6\%) & 30153 (63.1\%) & 1416 (3.0\%) & 4881 (10.2\%) & 47748 \\
 & totals & 56127 (23.8\%) & 146085 (61.9\%) & 8112 (3.4\%) & 25095 (10.6\%) & 235419 \\
\addlinespace
\bottomrule
\end{tabular}
\end{adjustbox}
\end{table}

Sleep staging (Number Breakdowns in Tables \ref{sleep})  is evaluated on the DREAMT dataset \citep{wang2024dreamt} hosted on PhysioNet \citep{goldberger2000physiobank}, which includes overnight wristband data with simultaneous PSG. Annotations follow AASM standards into wake, REM, NREM1, NREM2, and NREM3, excluding missing and preparation segments. PPG is bandpass filtered (0.5–12 Hz) \citep{butterworth1930theory}, segmented into 10s windows, and normalized to zero mean and unit variance. Performance is measured with five-fold subject-independent cross-validation. This task examines whether PPG encodes temporal patterns sufficient for sleep stage classification. \textit{We note that sleep staging has canonically been designed by leveraging the whole sleep cycle but we are assessing the ability to monitor real time sleep staging from much shorter PPG segments.} 

\subsubsection{Abnormal Lab Tests}

\begin{table}[h!]
\centering
\caption{Stratified 80/20 Train/Test splits for Tulane Abnormal Labs tasks (with per-task totals).}
\begin{adjustbox}{max width=0.6\linewidth}
\begin{tabular}{llrrr}
\toprule
Task & Split & Negative & Positive & Total \\
\midrule
\multirow{3}{*}{A1C} & train & 255 (31.6\%) & 553 (68.4\%) & 808 \\
 & test & 64 (31.7\%) & 138 (68.3\%) & 202 \\
 & totals & 319  & 691 & 1010 \\
\midrule
\multirow{3}{*}{Hematocrit} & train & 1271 (77.0\%) & 380 (23.0\%) & 1651 \\
 & test & 305 (77.0\%) & 91 (23.0\%) & 396 \\
 & totals & 1576 & 471 & 2047 \\
\midrule
\multirow{3}{*}{Hemoglobin} & train & 867 (81.2\%) & 201 (18.8\%) & 1068 \\
 & test & 208 (81.3\%) & 48 (18.8\%) & 256 \\
 & totals & 1075 & 249 & 1324 \\
\midrule
\multirow{3}{*}{Platelets} & train & 622 (35.5\%) & 1129 (64.5\%) & 1751 \\
 & test & 143 (35.7\%) & 258 (64.3\%) & 401 \\
 & totals & 765 & 1387 & 2152 \\
\midrule
\multirow{3}{*}{Potassium} & train & 731 (33.1\%) & 1476 (66.9\%) & 2207 \\
 & test & 167 (33.1\%) & 338 (66.9\%) & 505 \\
 & totals & 898 & 1814 & 2712 \\
\midrule
\multirow{3}{*}{Sodium} & train & 203 (17.6\%) & 951 (82.4\%) & 1154 \\
 & test & 48 (17.7\%) & 223 (82.3\%) & 271 \\
 & totals & 251 & 1174 & 1425 \\
\midrule
\multirow{3}{*}{WBC} & train & 247 (18.6\%) & 1082 (81.4\%) & 1329 \\
 & test & 62 (18.7\%) & 270 (81.3\%) & 332 \\
 & totals & 309 & 1352 & 1661 \\
\addlinespace
\bottomrule
\end{tabular}
\end{adjustbox}
\label{tulane}
\end{table}

For abnormal lab test prediction (Number Breakdowns in Tables \ref{tulane}), we use Samsung Watch PPG collected at Tulane University paired with clinical laboratory results. Each test is framed as a binary classification task: outcomes are labeled negative if within the 25th percentile of lab values and the positive labels are anything above the 75th percentile. All other labels are excluded. Preprocessing matches other tasks. Targets include A1C, hemoglobin, hematocrit, platelets, potassium, sodium, and WBC, each selected for established clinical relevance. This task extends evaluation beyond cardiovascular and behavioral endpoints to systemic markers of metabolic, renal, and hematologic health. \textit{We note that it is unclear whether PPG can predict abnormal from healthy lab values based on the PPG alone. Despite this, Tulane univeristy presents us with an opportunity to discover if PPG signal can provide digital signatures making this an exploratory task in our benchmark.}

\underline{Clinical Relevance of Lab Tests}
Each lab test used for this analysis provides critical information about a patient's health status. Their inclusion in this study is based on their established role in diagnosing or monitoring chronic conditions and acute health issues.

\begin{itemize}
    \item \textbf{A1C (Glycated Hemoglobin):} Measures average blood glucose levels over the past 2--3 months. It is the primary diagnostic tool for diabetes and a key indicator for managing long-term blood sugar control. Elevated A1C levels are linked to increased risk of cardiovascular disease, kidney damage, and other complications.
    \item \textbf{Hemoglobin:} Oxygen-carrying protein in red blood cells. Low levels indicate anemia, while elevated levels may suggest polycythemia vera.
    \item \textbf{Hematocrit:} Percentage of blood volume occupied by red blood cells. Used alongside hemoglobin to assess anemia or polycythemia.
    \item \textbf{Platelets:} Critical for clotting. Low count (thrombocytopenia) increases bleeding risk; high count (thrombocytosis) increases clot risk.
    \item \textbf{Potassium:} Essential electrolyte for nerve and muscle function. Both hypokalemia ($<$3.5 mEq/L) and hyperkalemia ($>$5.0 mEq/L) can trigger cardiac arrhythmias.
    \item \textbf{Sodium:} Regulates fluid balance and blood pressure. Abnormalities can indicate dehydration, renal disease, or endocrine disorders.
    \item \textbf{WBC (White Blood Cells):} Immune system cells. Leukocytosis ($>$11$\times$10\textsuperscript{9}/L) indicates infection, inflammation, or hematologic disease.
\end{itemize}

\clearpage
\section{Baselines and Model Configuration}
\label{baselines}

Self Supervised methods have become a dominant paradigm for health to study a variety of applications \citep{wornow2023shaky, lee2024feet, thieme2023foundation, he2024foundation, ono2024text, an2025raptor, lin2025case, thukral2025wavelet, lee2025towards, lee2024can}. Foundation models for one-dimensional signals are predominantly repurposed from architectures designed for vision, with adaptations that reinterpret temporal structure as a flattened analogue of spatial correlation. In this section we describe our baseline models and configurations 

\subsection{Baselines}

\textbf{LSM} \citep{narayanswamy2024scaling} introduces a large-scale foundation model trained on multimodal wearable sensor data. The approach adopts a vision transformer architecture trained via masked autoencoding with random masking. The model is designed as a general-purpose foundation, transferring effectively across a range of downstream tasks in physiological sensing and human activity recognition. In our work, we do not replicate the full multimodal design; instead, we adapt and constrain the model to a unimodal setting.

\textbf{Swin-Transformer} \citep{liu2021swin} is a hierarchical Transformer that forms multi-scale representations by restricting self-attention to non-overlapping windows and alternating partitions with a shifted-window scheme, which enables cross-window communication while keeping computation near-linear in sequence length. We use this baseline as this is a direct comparison and counterpart to our proposed hierarchical HiMAE model. For wearable sensing, we adopt a 1D adaptation that tokenizes temporal patches and applies windowed attention along time, capturing both fine-grained waveform morphology and longer-range dependencies. 

\textbf{Masked Siamese Networks} (MSN) \citep{assran2022masked} learn label-efficient representations by combining masked signal modeling with Siamese-style contrastive objectives. Instead of relying on class labels, MSN masks portions of the input and enforces consistency between augmented views. Architecturally, it employs a Vision Transformer encoder shared across views, while leveraging a predictor network to stabilize training. The key idea is to couple self-distillation with masked reconstruction to reduce sample complexity.

\textbf{DINO}  \citep{caron2021emergingpropertiesselfsupervisedvision} is a self-supervised framework that leverages knowledge distillation without labels. Using a teacher-student setup, the student network is trained to match the output distribution of the teacher under different data augmentations. Both networks are 1D-ViTs, and the method induces cluster-like emergent properties in the learned embedding space, enabling strong transfer performance without explicit contrastive pairs or handcrafted pretext tasks.

\textbf{SimCLR} \citep{chen2020simpleframeworkcontrastivelearning} establishes contrastive learning as a competitive self-supervised paradigm. The core idea is to maximize agreement between augmented views of the same signal in a latent space while pushing apart representations of different images. This is implemented using a ResNET encoder \citep{he2015deepresiduallearningimage}, a projection head, and a contrastive loss (NT-Xent \citep{chen2020simple}). 

\textbf{PaPaGei} \citep{pillai2024papagei} is a domain-specific foundation model designed for optical physiological sensing, particularly photoplethysmography (PPG). It adapts ResNET-style CNN architectures to learn robust, generalizable representations from large-scale optical physiological datasets. PaPaGei releases both model weights and datasets to support reproducibility and broader adoption in physiological signal analysis. In our work, we used their source code to benchmark their method by pre-training on our volume of data to ensure fair comparison.

\newpage
\subsection{Hyperparameters for HiMAE and Baselines}

To ensure a fair comparison across models, we aligned the training setup as closely as possible to the original implementations while maintaining consistency in optimizer choice and scheduling. All the methods trained from scratch (HiMAE, LSM, Swin-Transformer, MSN, DINO, SimCLR) were trained under identical optimization regimes, while PaPaGei follows its released open source training protocol. Table~\ref{modelconfigs} summarizes the key hyperparameters for all models.

\begin{table}[th!]
\centering
\caption{Hyperparameter Configurations for Different Models}
\label{modelconfigs}
\begin{adjustbox}{max width=\textwidth}
\begin{tabular}{l|ccccccc}
\toprule
\textbf{Configuration}
& \textbf{HiMAE} & \textbf{LSM} & \textbf{Swin-Transformer} & \textbf{MSN} & \textbf{DINO} & \textbf{SimCLR} & \textbf{PaPaGei} \\
\midrule
\textbf{Training Steps} & \multicolumn{6}{c}{100000} & 15000 \\
\textbf{Warmup Steps} & \multicolumn{6}{c}{2500} & --- \\
\textbf{Optimizer} & \multicolumn{7}{c}{AdamW (\cite{loshchilov2019decoupled})} \\
\textbf{Opt. momentum [$\beta_1, \beta_2$]}
& [0.9, 0.95] & [0.9, 0.95] & [0.9, 0.95] & [0.9, 0.99] & [0.9, 0.99] & [0.9, 0.99] & --- \\
\textbf{Base learning rate} &
0.001 & 0.005 & 0.005 & 0.001 & 0.004 & 0.001 & 0.0001 \\
\textbf{Batch size} & \multicolumn{6}{c}{2048} & 256 \\
\textbf{Weight decay} & \multicolumn{6}{c}{0.0001} & --- \\
\textbf{Gradient clipping} & 1.0 & 1.0 & 1.0 & 3.0 & 3.0 & 3.0 & --- \\
\textbf{Dropout} & \multicolumn{6}{c}{0.0} & --- \\
\textbf{Learning rate schedule} & \multicolumn{6}{c}{Linear Warmup \& Cosine Decay} & --- \\
\textbf{Loss Function} & \multicolumn{4}{c}{Mean Squared Error} & Cross Entropy & \multicolumn{2}{c}{Contrastive Loss} \\
\textbf{Data resolution} & \multicolumn{7}{c}{1 (signal) - 100 Hz (Sampling Rate) $\times$ 10 (seconds)} \\
\textbf{Augmentation} & \multicolumn{7}{c}{Flip, Time-Warping, Noise} \\
\bottomrule
\end{tabular}
\end{adjustbox}
\end{table}

\newpage

\section{Additional Results}
\label{addresults}

\subsection{Model Configurations Ablations}
\label{ablation-hyper}

We conducted a comprehensive ablation study of HiMAE on a 100 Hz dataset comprising ten million segments (roughly 30k hours). The experiments systematically varied architecture and hyperparameters to understand their effect on reconstruction quality (Extrapolation task from our generative benchmark in tables where it is not explicitly stated as previously done in \citep{narayanswamy2024scaling}), with multiple independent training runs averaged to reduce variance from stochastic initialization and data sampling. Unless otherwise noted, all training employed AdamW with a learning rate of $3\times10^{-4}$, cosine decay scheduling, and a batch size of 512.

\noindent\underline{Architecture.}  

We evaluated HiMAE alongside CNN baselines across increasing network depths, defined by the sequence of hidden channel dimensions $[16,32,64]$, $[16,32,64,128]$, and $[16,32,64,128,256]$. Table~\ref{tab:params} lists the parameter counts, showing a modest growth for HiMAE compared to CNN baselines, with the skip-connected HiMAE exhibiting slightly higher capacity than its no-skip variant.

\begin{table}[h!]
    \centering
    \caption{Model Parameters (in K or M)}
    \label{tab:params}
    \begin{tabular}{lrrr}
        \toprule
        \textbf{Model} & \textbf{HiMAE-tiny} & \textbf{HiMAE-small} & \textbf{HiMAE-Base} \\
        \textit{Depth} & [16,32,64] & [16,32,64,128] & [16,32,64,128,256] \\
        \midrule
        CNN & 26.2 K & 108 K & 437 K \\
        HiMAE-no skip & 66.1 K & 271 K & 1.10 M \\
        HiMAE & 75.3 K & 309 K & 1.25 M \\
        \bottomrule
    \end{tabular}
\end{table}

The impact of network depth on mean absolute error (MAE) and mean squared error (MSE) is summarized in Table~\ref{tab:depth}. Increasing depth consistently reduced both MAE and MSE for HiMAE, with the deepest configuration yielding the lowest reconstruction error. Skip connections were critical, as HiMAE consistently outperformed its no-skip variant across all depths.

\begin{table}[htbp]
    \centering
    \caption{MAE and MSE for Different Network Depths}
    \label{tab:depth}
    \begin{tabular}{@{}l*{6}{S[table-format=1.4]}@{}}
        \toprule
                \textbf{Model} & \multicolumn{2}{c}{\textbf{HiMAE-tiny}} & \multicolumn{2}{c}{\textbf{HiMAE-small}} & \multicolumn{2}{c}{\textbf{HiMAE-Base}} \\
        \multirow{2}{*}{\textit{Depth}} & \multicolumn{2}{c}{\text{[16,32,64]}} & \multicolumn{2}{c}{\text{[16,32,64,128]}} & \multicolumn{2}{c}{\text{[16,32,64,128,256]}} \\
        \cmidrule(lr){2-3} \cmidrule(lr){4-5} \cmidrule(lr){6-7}
        & \multicolumn{1}{c}{\textbf{MAE} $\downarrow$} & \multicolumn{1}{c}{\textbf{MSE} $\downarrow$} & \multicolumn{1}{c}{\textbf{MAE} $\downarrow$} & \multicolumn{1}{c}{\textbf{MSE} $\downarrow$} & \multicolumn{1}{c}{\textbf{MAE} $\downarrow$} & \multicolumn{1}{c}{\textbf{MSE} $\downarrow$} \\
        \midrule
        CNN & 0.4052 & 0.2345 & 0.4177 & 0.2491 & 0.4008 & 0.2315 \\
        HiMAE-noskip & 0.4031 & 0.2365 & 0.4006 & 0.2465 & 0.3975 & 0.2339 \\
        HiMAE & 0.4008 & 0.2309 & 0.3892 & 0.2232 & \textbf{0.3827} & \textbf{0.2210} \\
        \bottomrule
    \end{tabular}
\end{table}



\noindent\underline{Patch Size.}  

We varied the spatial-temporal patch sizes over $1$, $5$, $10$, and $20$. The results in Table~\ref{tab:patchsize} indicate that $5$ provided the best trade-off between local resolution and generative performance. Smaller patches increased flexibility but slightly degraded performance due to reduced receptive field per token, while overly large patches caused loss of fine-grained structure.

\begin{table}[htbp]
    \centering
    \caption{Model Performance for Different Patch Sizes}
    \label{tab:patchsize}
    \begin{tabular}{@{}l*{8}{S[table-format=1.4]}@{}}
        \toprule
        \multirow{2}{*}{\textbf{Model}} & \multicolumn{2}{c}{\textbf{1}} & \multicolumn{2}{c}{\textbf{5}} & \multicolumn{2}{c}{\textbf{10}} & \multicolumn{2}{c}{\textbf{20}} \\
        \cmidrule(lr){2-3} \cmidrule(lr){4-5} \cmidrule(lr){6-7} \cmidrule(lr){8-9}
        & \multicolumn{1}{c}{\textbf{MAE} $\downarrow$} & \multicolumn{1}{c}{\textbf{MSE} $\downarrow$} & \multicolumn{1}{c}{\textbf{MAE} $\downarrow$} & \multicolumn{1}{c}{\textbf{MSE} $\downarrow$} & \multicolumn{1}{c}{\textbf{MAE} $\downarrow$} & \multicolumn{1}{c}{\textbf{MSE} $\downarrow$} & \multicolumn{1}{c}{\textbf{MAE} $\downarrow$} & \multicolumn{1}{c}{\textbf{MSE} $\downarrow$} \\
        \midrule
        CNN & 0.4140 & 0.2391 & 0.4008 & 0.2315 & 0.4122 & 0.2449 & 0.4274 & 0.2613 \\
        HiMAE-noskip & 0.4069 & 0.2398 & 0.3976 & 0.2339 & 0.4037 & 0.2462 & 0.4195 & 0.2629 \\
        HiMAE & 0.3899 & 0.2268 & \textbf{0.3827} & \textbf{0.2210} & 0.3861 & 0.2312 & 0.4039 & 0.2479 \\
        \bottomrule
    \end{tabular}
\end{table}

\noindent\underline{Convolution Kernel Size.}  

Kernel size was varied over $\{1,5,10,20\}$. Table~\ref{tab:kernelsize} shows that $5$ yielded the lowest errors across all models, suggesting moderate receptive fields match the temporal and spatial scales of our data. Very small kernels restricted context aggregation, while very large kernels oversmoothed latent features.

\begin{table}[htbp]
    \centering
    \caption{Model Performance Across Convolution Kernel Sizes}
    \label{tab:kernelsize}
    \begin{tabular}{@{}l*{8}{S[table-format=1.4]}@{}}
        \toprule
        \multirow{2}{*}{\textbf{Model}} & \multicolumn{2}{c}{\textbf{1}} & \multicolumn{2}{c}{\textbf{5}} & \multicolumn{2}{c}{\textbf{10}} & \multicolumn{2}{c}{\textbf{20}} \\
        \cmidrule(lr){2-3} \cmidrule(lr){4-5} \cmidrule(lr){6-7} \cmidrule(lr){8-9}
        & \multicolumn{1}{c}{\textbf{MAE} $\downarrow$} & \multicolumn{1}{c}{\textbf{MSE} $\downarrow$} & \multicolumn{1}{c}{\textbf{MAE} $\downarrow$} & \multicolumn{1}{c}{\textbf{MSE} $\downarrow$} & \multicolumn{1}{c}{\textbf{MAE} $\downarrow$} & \multicolumn{1}{c}{\textbf{MSE} $\downarrow$} & \multicolumn{1}{c}{\textbf{MAE} $\downarrow$} & \multicolumn{1}{c}{\textbf{MSE} $\downarrow$} \\
        \midrule
        CNN & 0.4162 & 0.2413 & 0.4010 & 0.2309 & 0.4103 & 0.2418 & 0.4241 & 0.2576 \\
        HiMAE-noskip & 0.4090 & 0.2427 & 0.3959 & 0.2331 & 0.4032 & 0.2440 & 0.4208 & 0.2591 \\
        HiMAE & 0.3921 & 0.2283 & \textbf{0.3821} & \textbf{0.2206} & 0.3885 & 0.2316 & 0.4047 & 0.2485 \\
        \bottomrule
    \end{tabular}
\end{table}

\noindent\underline{Stride.}  

We evaluated stride values of $2$, $4$, and $8$ (Table~\ref{tab:stride}). Smaller strides yielded the best performance, particularly for HiMAE, by preserving high temporal resolution in early feature maps. Performance degraded monotonically with stride increases.

\begin{table}[htbp]
    \centering
    \caption{Model Performance Across Stride Values}
    \label{tab:stride}
    \begin{tabular}{@{}l*{6}{S[table-format=1.4]}@{}}
        \toprule
        \multirow{2}{*}{\textbf{Model}} & \multicolumn{2}{c}{\textbf{2}} & \multicolumn{2}{c}{\textbf{4}} & \multicolumn{2}{c}{\textbf{8}} \\
        \cmidrule(lr){2-3} \cmidrule(lr){4-5} \cmidrule(lr){6-7}
        & \multicolumn{1}{c}{\textbf{MAE} $\downarrow$} & \multicolumn{1}{c}{\textbf{MSE} $\downarrow$} & \multicolumn{1}{c}{\textbf{MAE} $\downarrow$} & \multicolumn{1}{c}{\textbf{MSE} $\downarrow$} & \multicolumn{1}{c}{\textbf{MAE} $\downarrow$} & \multicolumn{1}{c}{\textbf{MSE} $\downarrow$} \\
        \midrule
        CNN & \textbf{0.4016} & \textbf{0.2312} & 0.4139 & 0.2445 & 0.4318 & 0.2678 \\
        HiMAE-noskip & \textbf{0.3976} & \textbf{0.2334} & 0.4098 & 0.2471 & 0.4272 & 0.2702 \\
        HiMAE & \textbf{0.3829} & \textbf{0.2209} & 0.3928 & 0.2325 & 0.4103 & 0.2504 \\
        \bottomrule
    \end{tabular}
\end{table}

\noindent\underline{Masking Ratio.}  

Finally, we explored the effect of varying the latent masking ratio in the masked autoencoding objective for generative tasks, with ratios from $0.5$ to $0.9$. As shown in Table~\ref{tab:maskratio}, interpolation and extrapolation both improved when increasing the ratio up to $0.8$, after which performance degraded for interpolation and collapsed for extrapolation.

\begin{table}[htbp]
    \centering
    \caption{MAE and MSE for HiMAE Across Different Masking Ratios Evaluated on Generative Tasks}
    \label{tab:maskratio}
    \begin{tabular}{@{}l*{4}{S[table-format=1.4]}@{}}
        \toprule
        \multirow{2}{*}{\textbf{HiMAE Masking Ratio}} & \multicolumn{2}{c}{\textbf{Temporal Interpolation}} & \multicolumn{2}{c}{\textbf{Temporal Extrapolation}} \\
        \cmidrule(lr){2-3} \cmidrule(lr){4-5}
        & \multicolumn{1}{c}{\textbf{MAE} $\downarrow$} & \multicolumn{1}{c}{\textbf{MSE} $\downarrow$} & \multicolumn{1}{c}{\textbf{MAE} $\downarrow$} & \multicolumn{1}{c}{\textbf{MSE} $\downarrow$} \\
        \midrule
        0.5 & 0.3972 & 0.2292 & 0.4077 & 0.2519 \\
        0.6 & 0.3889 & 0.2223 & 0.3975 & 0.2294 \\
        0.7 & 0.3848 & 0.2207 & 0.3963 & 0.2278 \\
        \rowcolor{gray!15} 0.8 & \textbf{0.3796} & \textbf{0.2183} & \textbf{0.3879} & \textbf{0.2217} \\
        0.9 & 0.3818 & 0.2219 & 0.2881 & 0.2216 \\
        \bottomrule
    \end{tabular}
\end{table}

\noindent\underline{Final Selection.}  

These controlled experiments informed the final HiMAE configuration: the deepest architecture $[16,32,64,128,256]$ with skip connections, patch size $5$, kernel size $5$, stride $2$, and a masking ratio of $0.8$, which jointly achieved the best trade-off between reconstruction fidelity and parameter efficiency.

\subsection{ECG Pre-training}
\label{ecg-fm}

\begin{wraptable}[9]{r}{0.52\linewidth}
\centering
\vspace{-0.8em}
\caption{Masked-reconstruction loss on ECG masked auto encoding task.}
\label{ecg_loss}
\begin{tabular}{l c}
\toprule
Model & MSE ($\downarrow$) \\
\midrule
HiMAE & \textbf{0.148} \\
LSM-1 (ViT) & 0.162 \\
HiMAE (no skip) & 0.184 \\
CNN & 0.207 \\
\bottomrule
\end{tabular}
\vspace{-0.8em}
\end{wraptable}

HiMAE attains the lowest masked-reconstruction error on ECG (Table~\ref{ecg_loss}), indicating that its hierarchical masking and reconstruction inductive biases capture reconstruction capacity beyond PPG. LSM-1 (ViT) is a close second, while the ablated HiMAE and CNN trail, reinforcing that the full HiMAE design transfers effectively to the ECG domain.

\newpage
\subsection{Visualization of reconstructions}

\begin{figure}[h!]
    \centering
    \includegraphics[width=0.75\linewidth]{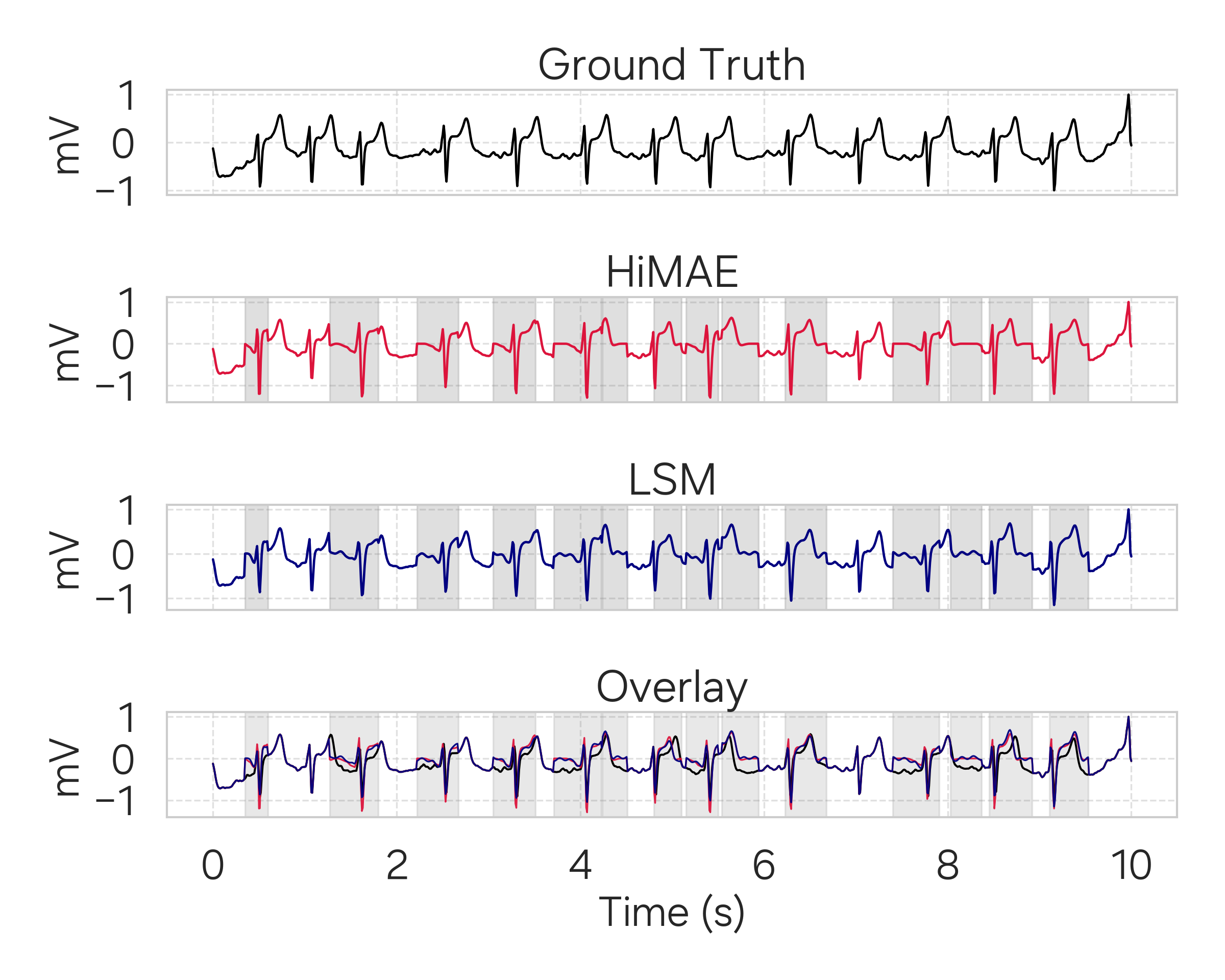}
    \vspace{-0.8cm}
    \includegraphics[width=0.75\linewidth]{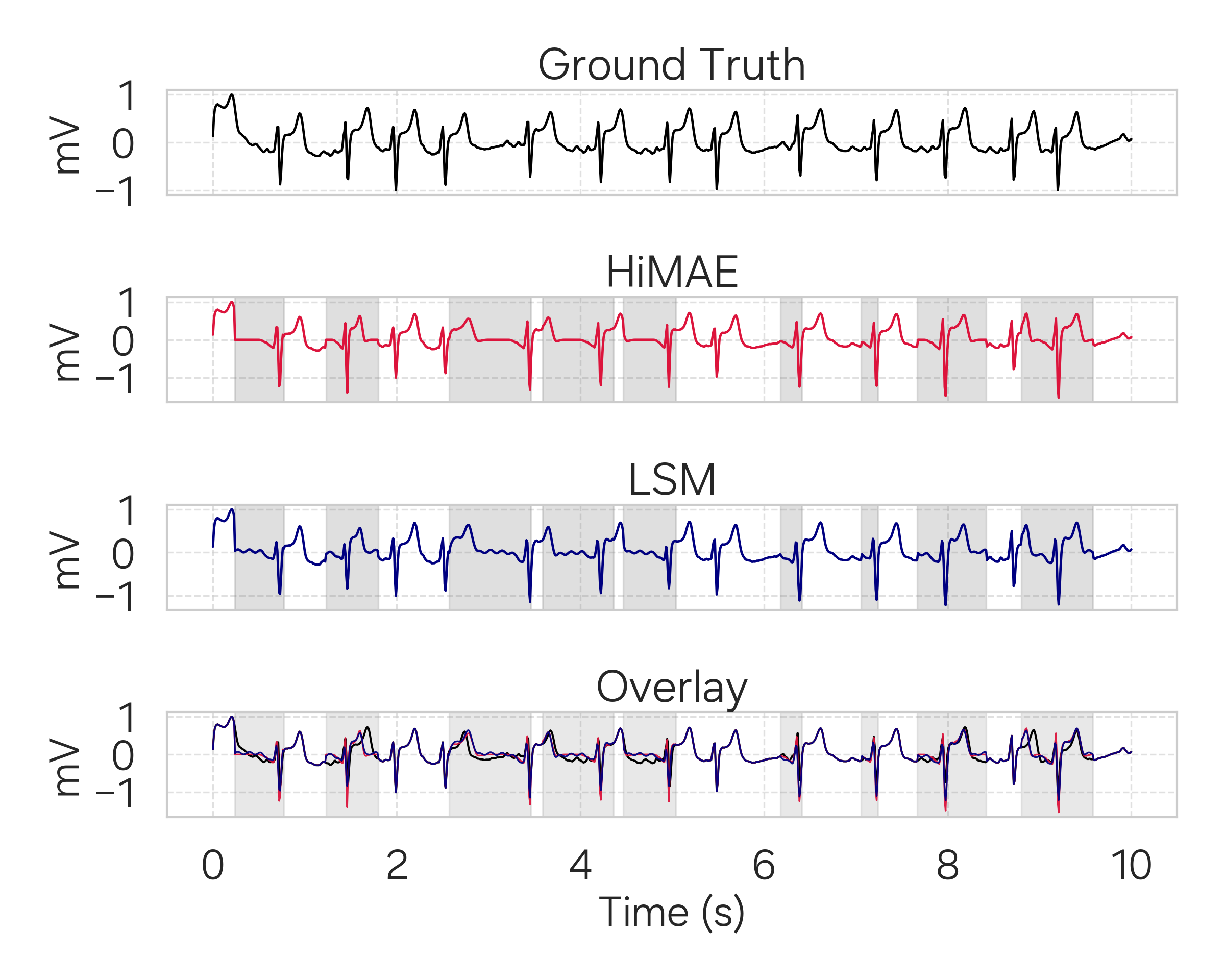}
    \caption{\textbf{ECG Reconstructions:} ECG Sample Reconstructions for HiMAE \& LSM}
    \label{ecg-reconstruction}
\end{figure}

\begin{figure}[h!]
    \centering
    \includegraphics[width=0.75\linewidth]{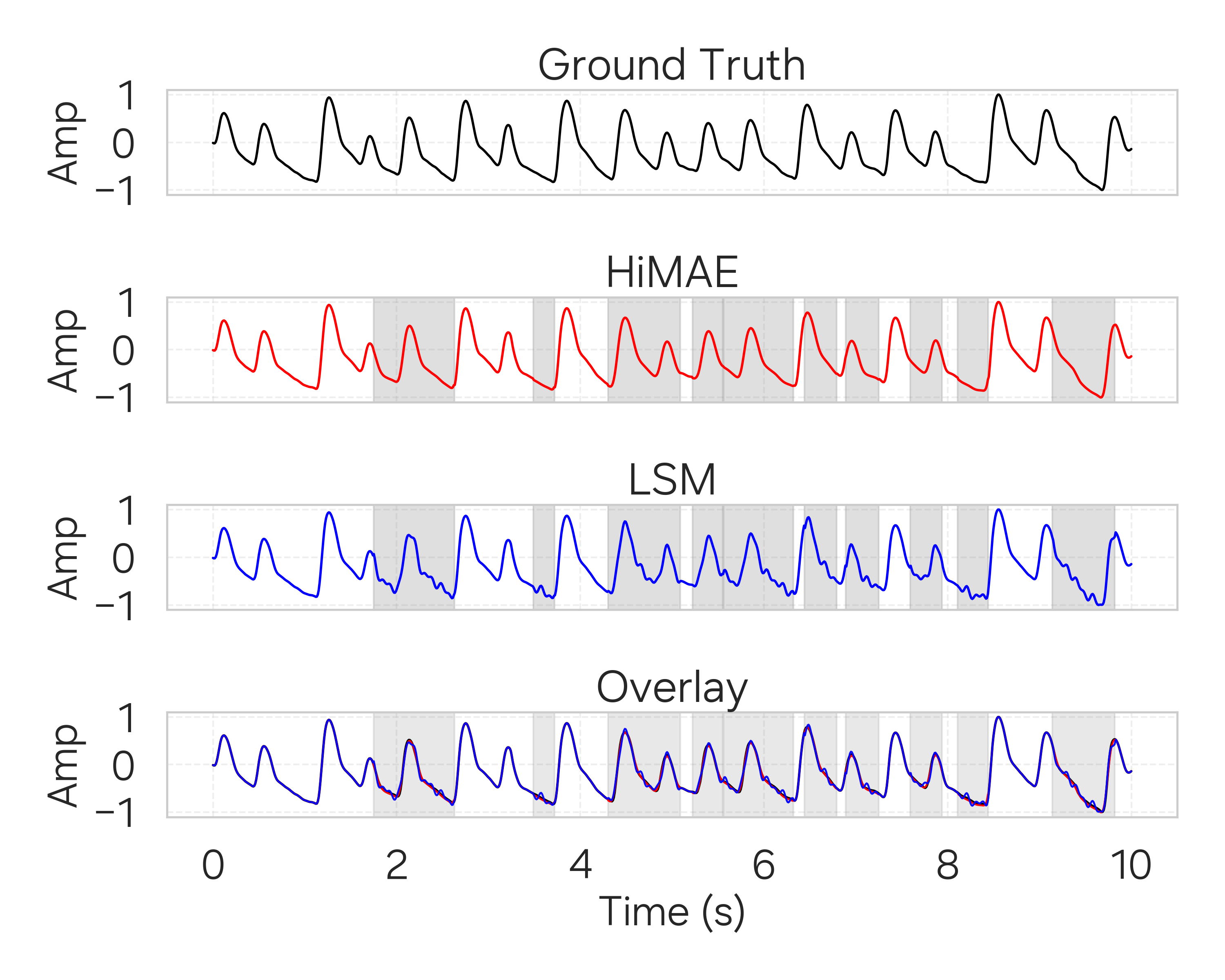}
    \vspace{-0.8cm}\includegraphics[width=0.75\linewidth]{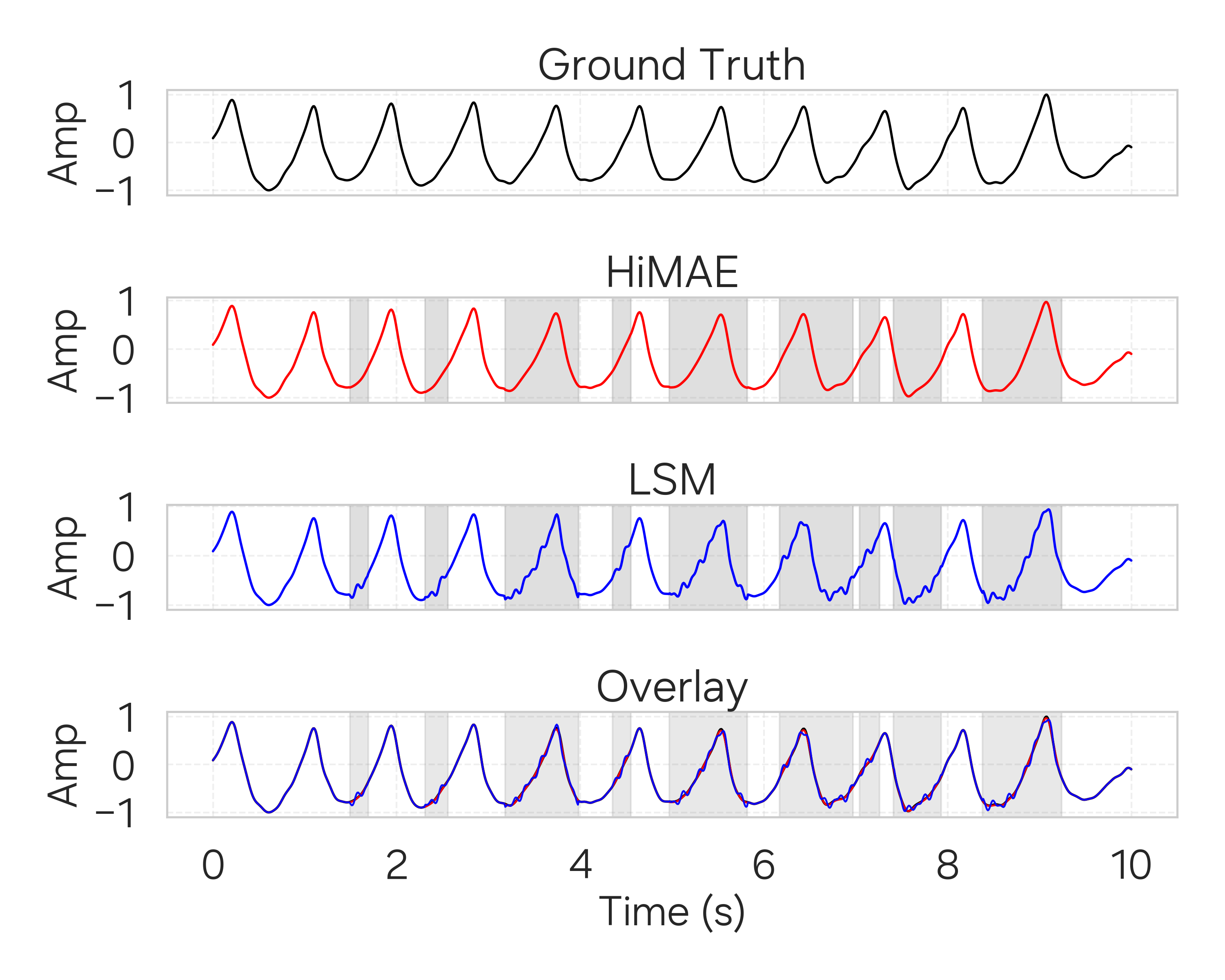}
    \caption{\textbf{PPG Reconstructions:} PPG Sample Reconstructions for HiMAE \& LSM}
    \label{ppg-reconstruction}
\end{figure}

We provide sample reconstructions on both ECG (Figure \ref{ecg-reconstruction}) and PPG (Figure \ref{ppg-reconstruction})  signal showcasing that our framework works across signal modalities. In our work, we limit our analysis to PPG, since ECG is not passively collected and obtaining paired PPG and ECG data was not attainable at scale.

\clearpage
\newpage 
\subsection{Scaling Results for Generative Tasks}
\label{scaling-generative}

\begin{figure}[h!]
    \centering
    \includegraphics[width=0.9\linewidth]{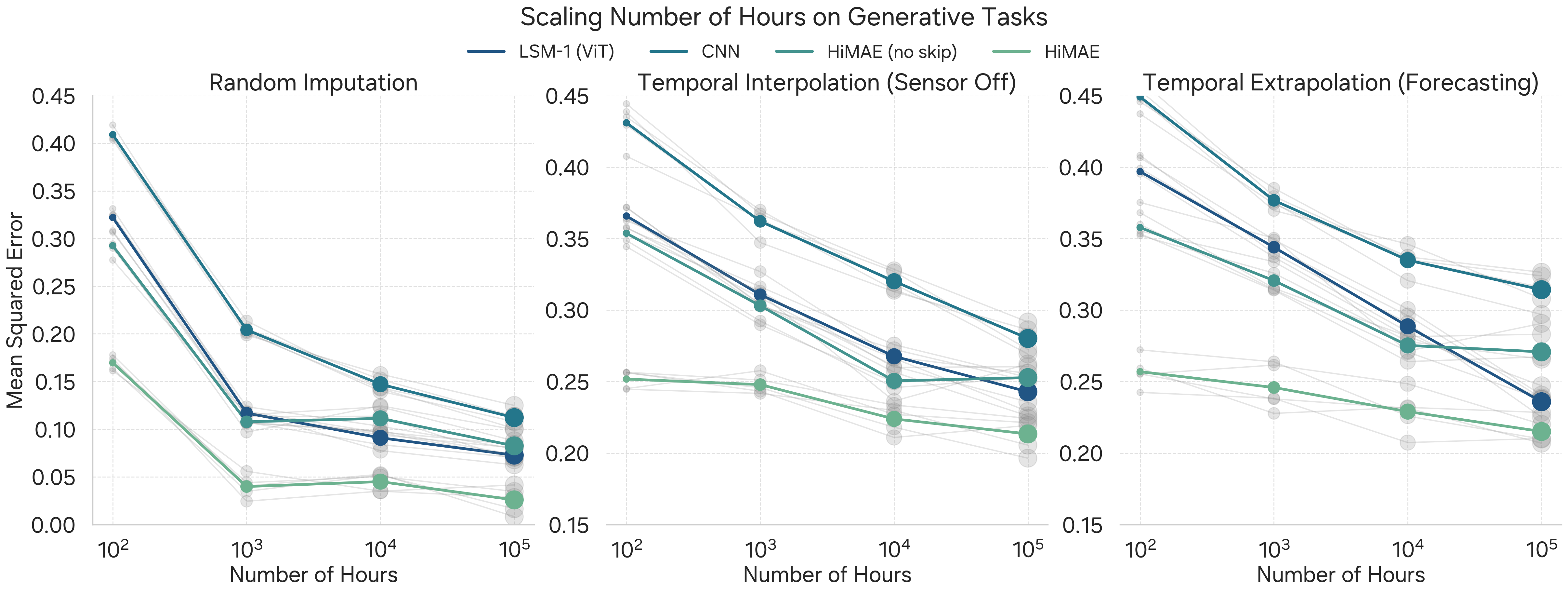}
    \includegraphics[width=0.9\linewidth]{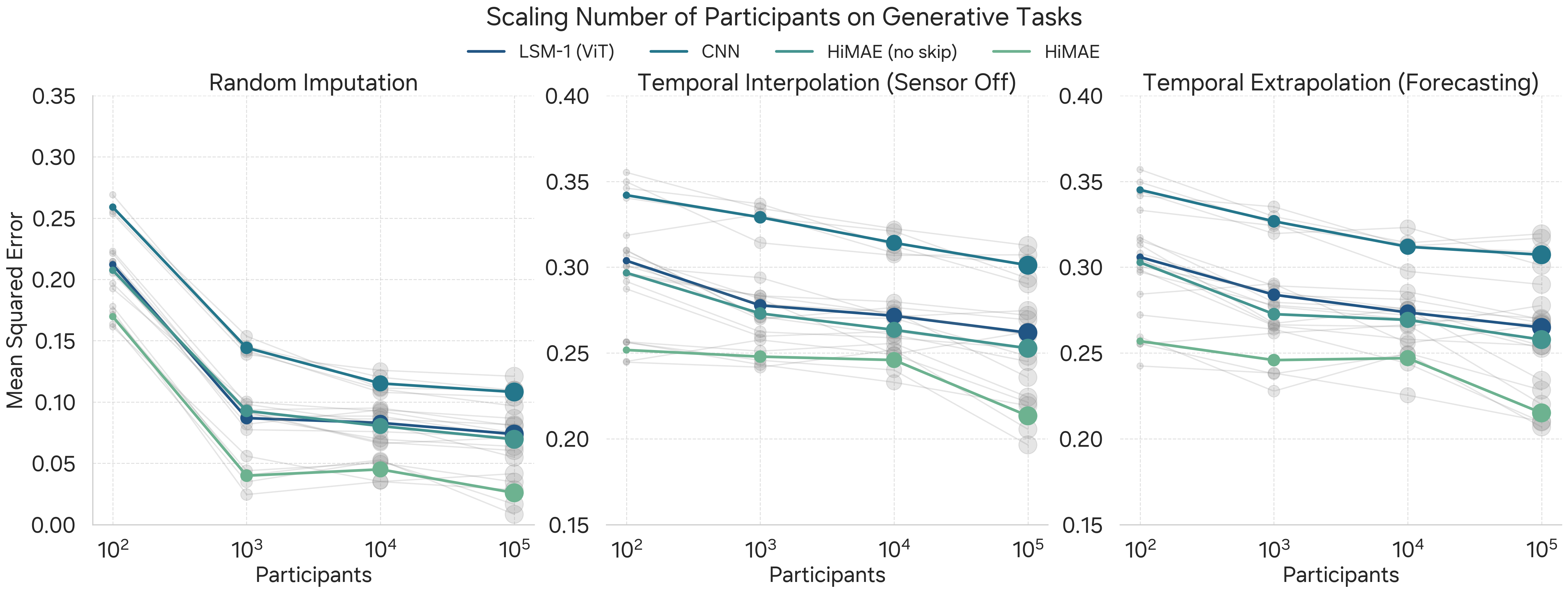}
    \includegraphics[width=0.9\linewidth]{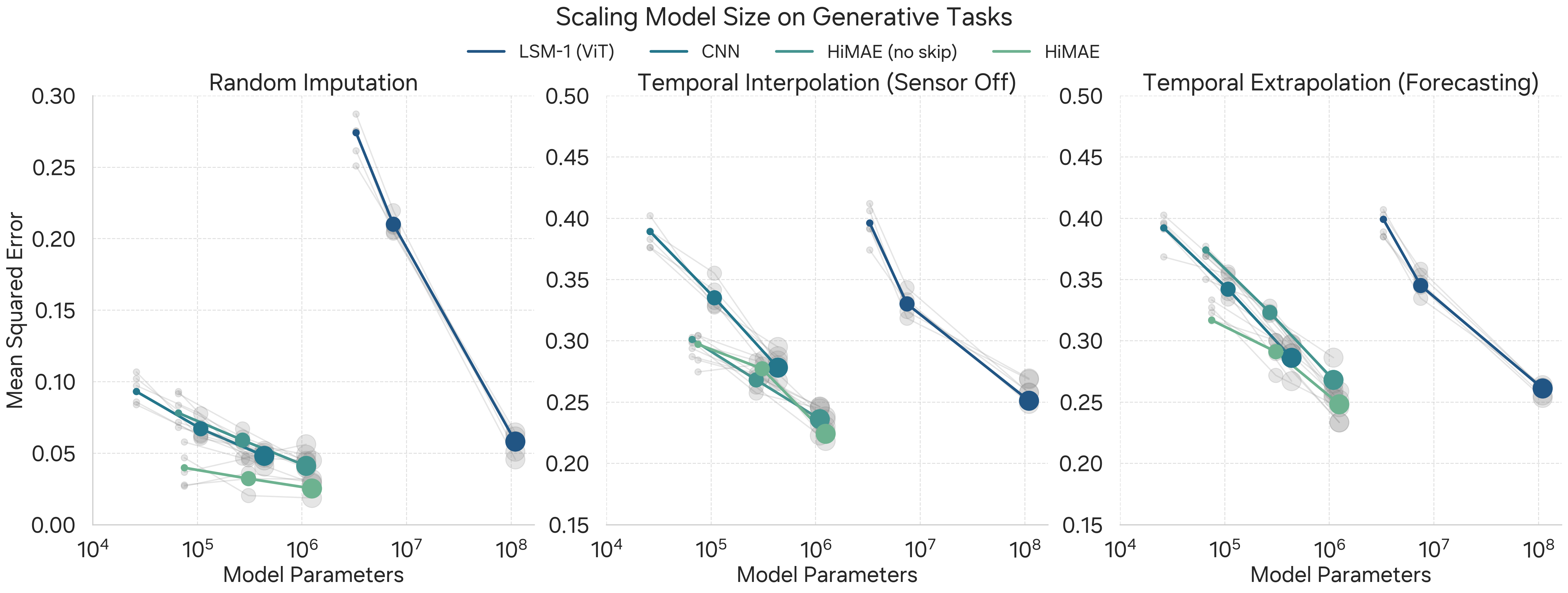}
    \includegraphics[width=0.9\linewidth]{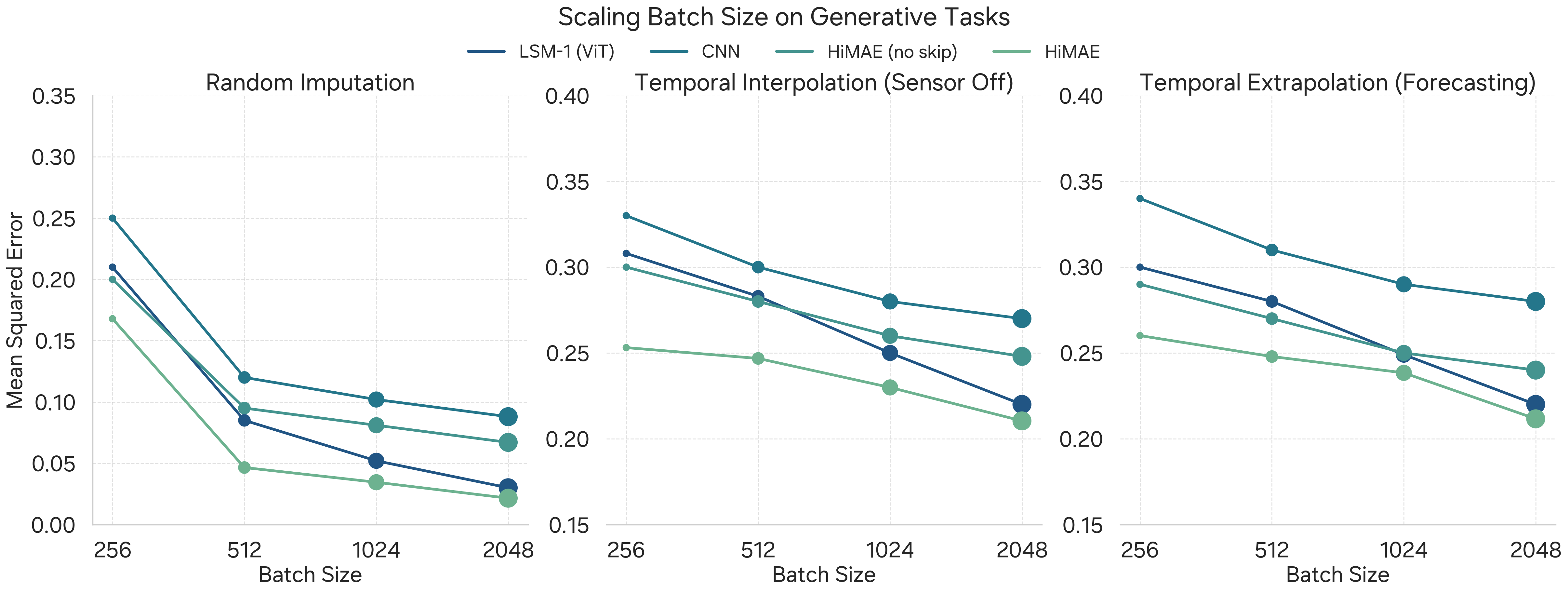}
    \caption{\textbf{Scaling Experiments on Generative Tasks:} Evaluation on the three generative tasks. HiMAE consistenly outperforms all model at our scale of data}
    \label{scaling2}
\end{figure}

\label{scaling-generative}
\underline{Scaling analysis.} We evaluate HiMAE’s reconstruction error under participant, recording hour, batch size, and model size scaling, following the regimes of \citet{narayanswamy2024scaling, xu2025lsm}: random imputation, temporal interpolation, and temporal extrapolation (Figure \ref{scaling2}). Across all settings HiMAE follows clean  scaling law trends \citep{kaplan2020scaling} and maintains a margin over LSM-1 (ViT) and CNN baselines.  

The most pronounced effect is model size. At small capacities HiMAE achieves lower error than much larger transformer baselines, highlighting the advantage of hierarchical inductive bias over sheer parameter count. LSM-1 only begins to close the gap at orders of magnitude more parameters. The transformer could surpass our HiMAE model when given a larger capacity but this again highlights the effectiveness of the inductive bias that we are conveying.  Participant, hour, and batch size scaling follow canonical patterns. More participants and longer recordings steadily reduce error, with HiMAE continuing to improve where baselines saturate, especially on interpolation and extrapolation. 

Ablations confirm the mechanism: removing skip connections or collapsing the hierarchy to a single scale uniformly degrades performance, with gaps widening as data or model size grow. Task difficulty follows the expected order (imputation $<$ interpolation $<$ extrapolation), with the largest relative gaps in extrapolation, where hierarchical structure effectively lengthens usable context. Overall, HiMAE reaches lower error at smaller scales, showing that efficiency derives from inductive bias rather than brute force capacity.

\subsection{Hierarchal Concordance}

\label{agreement}

\begin{figure}
    \centering
    \includegraphics[width=\linewidth]{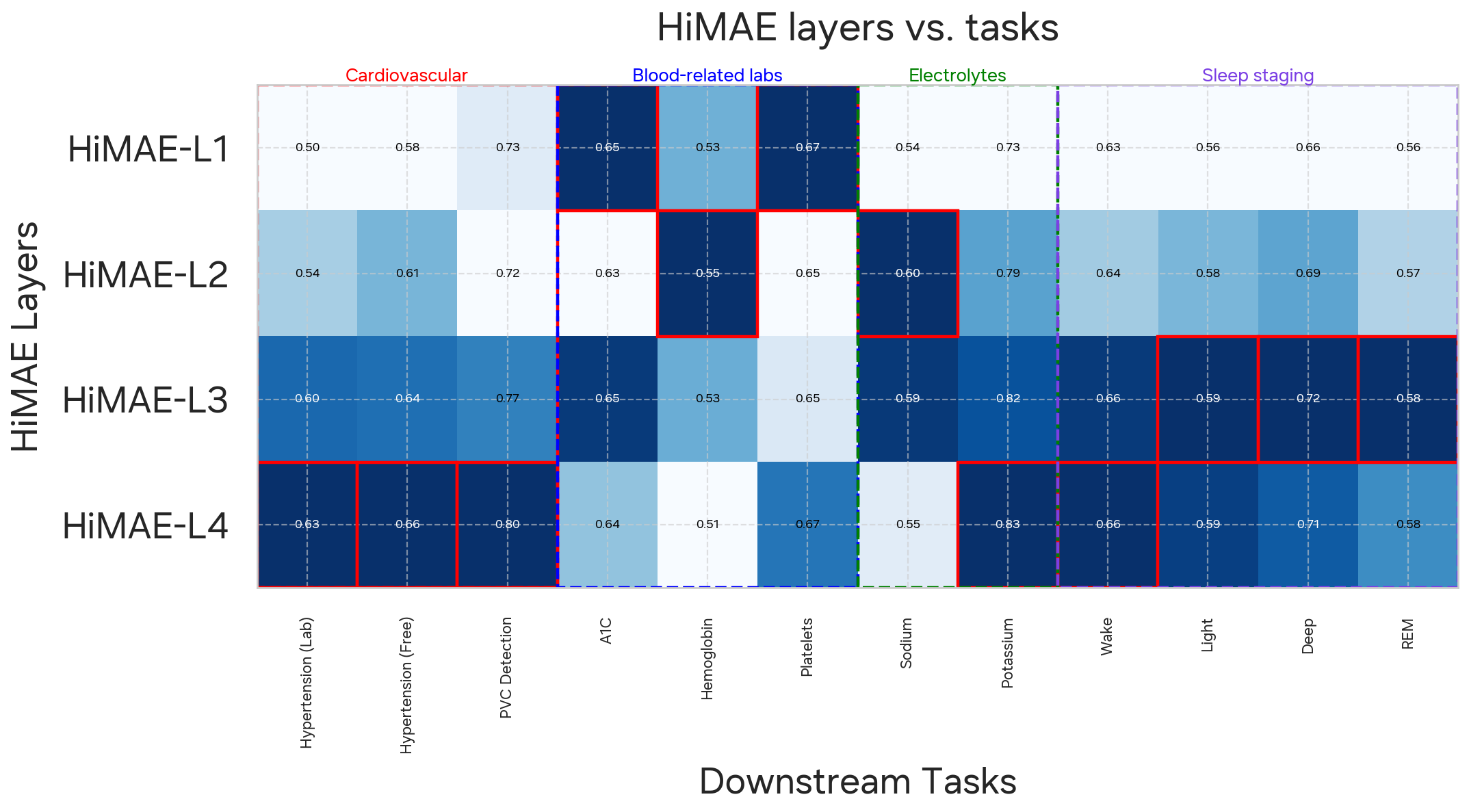}
    \includegraphics[width=\linewidth]{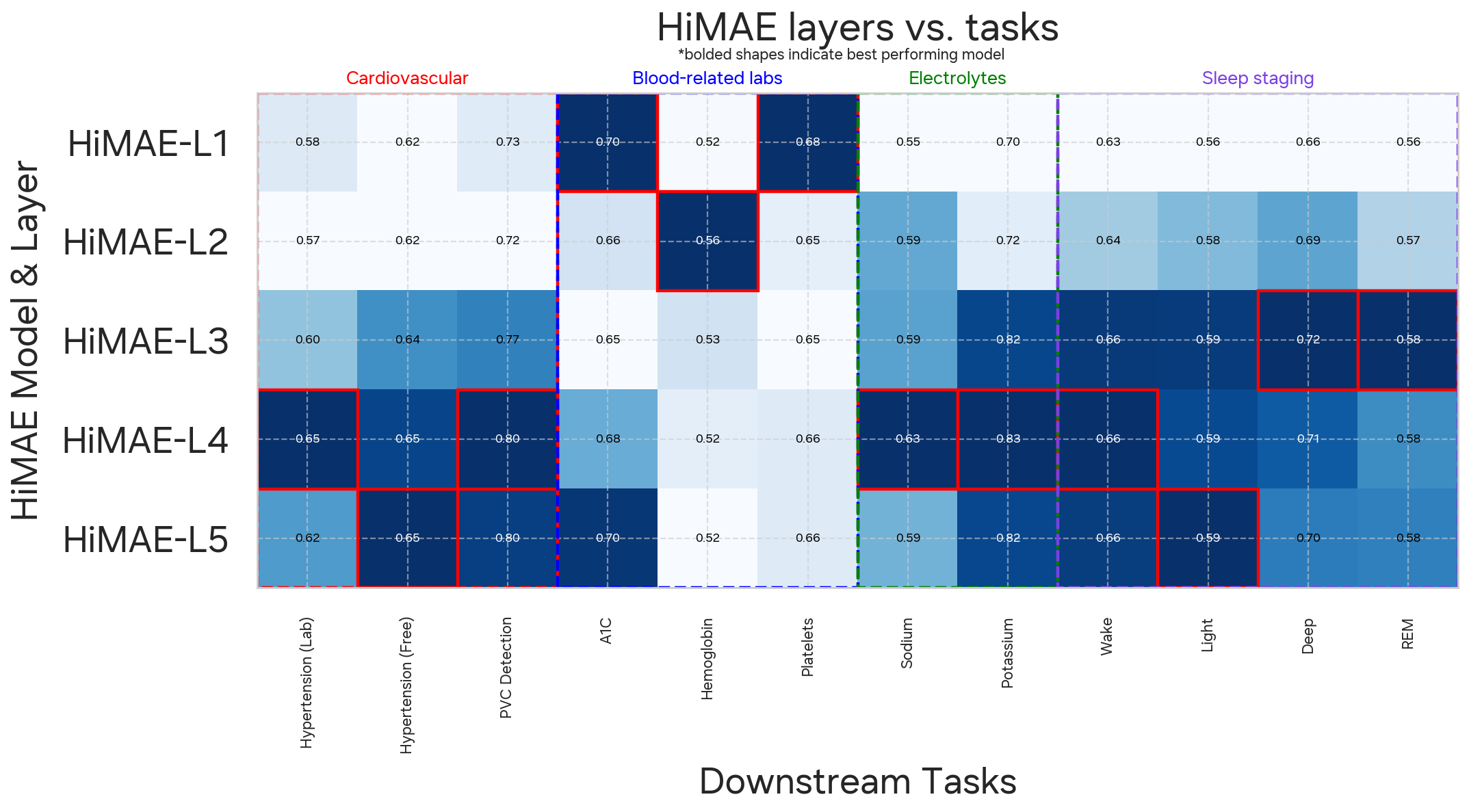}
    \caption{\textbf{HiMAE layer concordance across encoder depths.}  Heatmaps compare downstream AUROC when probing HiMAE at 4 layers (top) versus 5 layers (bottom). Despite the removal of an encoder–decoder stage, the resolution–task alignment remains highly concordant: tasks such as PVC detection and hypertension consistently peak at similar layers, while sleep staging benefits from coarser representations. Minor deviations appear in intermediate layers, but the overall hierarchy of predictive resolutions is preserved, indicating robustness of the resolution hypothesis to architectural depth.
}
    \label{concordance}
\end{figure}

\underline{Layer concordance across depths.}  
We further assess the stability of the resolution hypothesis by comparing HiMAE trained with four versus five encoder–decoder stages (Figure~\ref{concordance}). The resulting heatmaps reveal that the alignment between downstream tasks and temporal resolutions is largely preserved across depths. Cardiovascular endpoints such as PVC detection and hypertension consistently achieve their best performance at finer layers, while blood related labs benefits from coarser layers. Although minor fluctuations appear in intermediate levels, the overall hierarchy of predictive resolutions is concordant. This suggests that the resolution–task mapping uncovered by HiMAE is not an artifact of architectural depth, but a robust property of the representations themselves.

\subsection{Regression}
\label{regression-section}

\begin{figure*}
  \includegraphics[width=\textwidth]{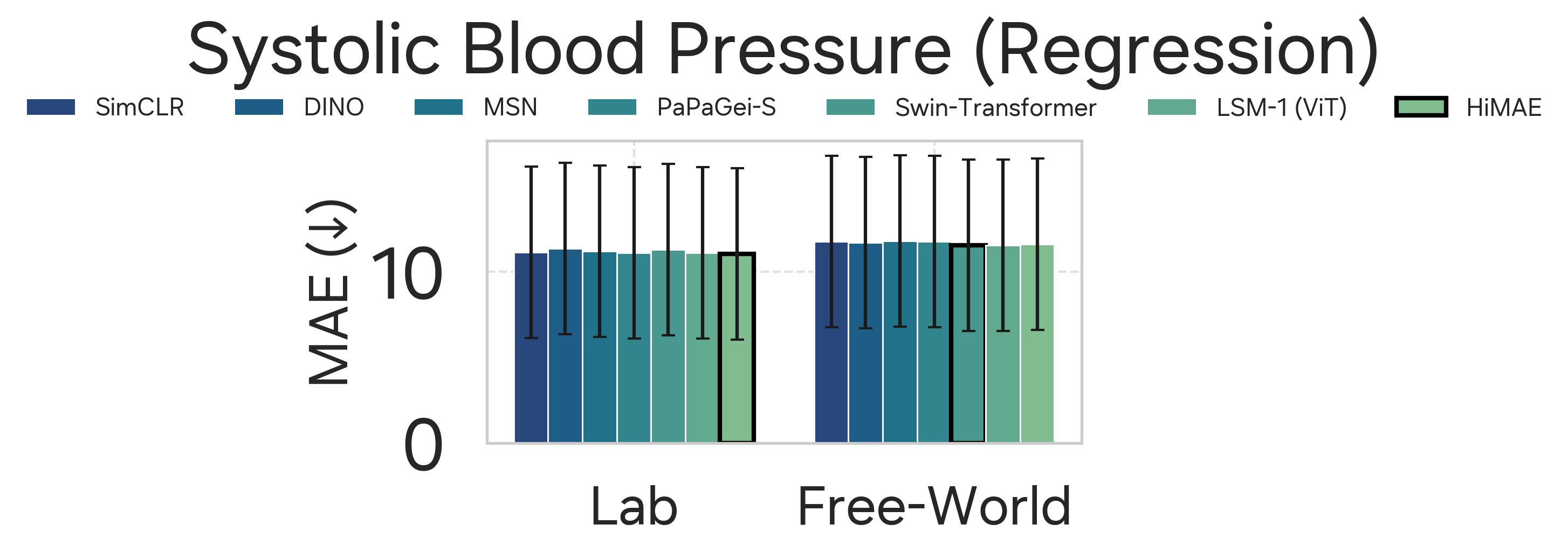}
  \includegraphics[width=\textwidth]{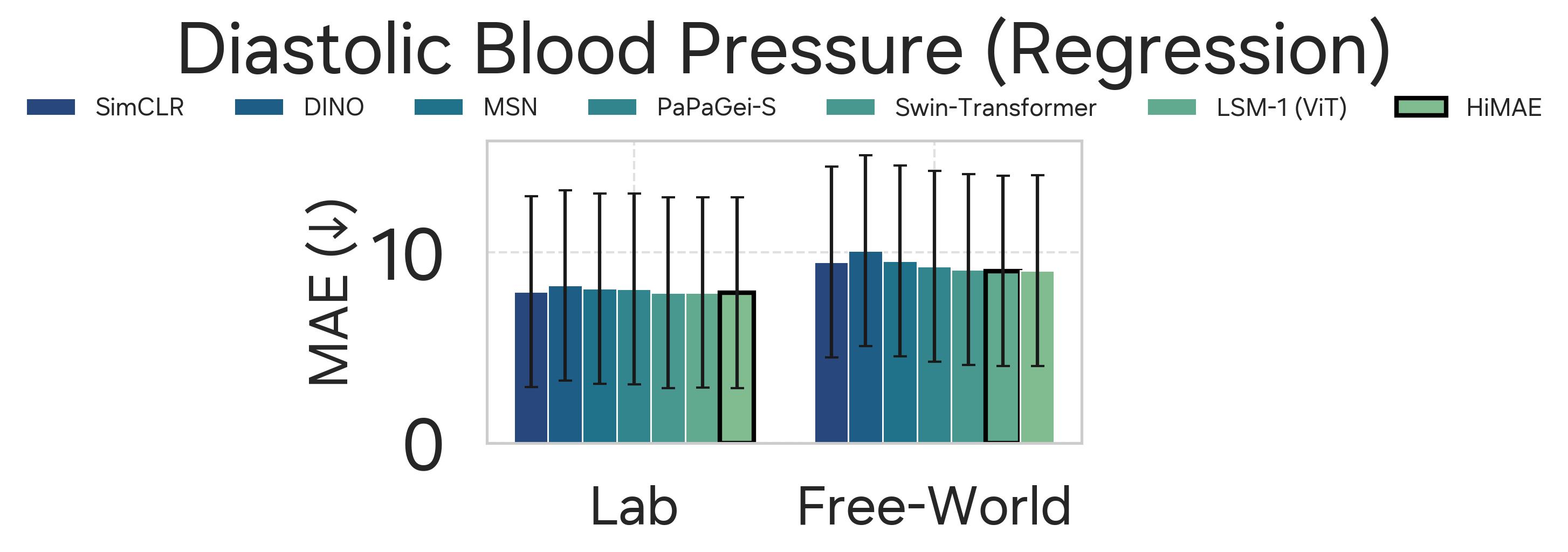}
  \caption{\textbf{Performance on regression benchmarks.} Mean absolute error ($\downarrow$) for regressing systolic and diastolic blood pressure.
}
  \label{regression}
\end{figure*}

Continuous regression of blood pressure from wearable signals represents a canonical benchmark for physiological monitoring, yet the task remains highly challenging \citep{schrumpf2021assessment, schrumpf2021assessment2, mehta2024examining}. The objective is to recover systolic and diastolic pressures directly from sensor data, a setting where accuracy demands are clinically stringent but input signals are noisy and weakly correlated with the target (Figure ~\ref{regression}). On the diastolic task, all approaches converge to errors on the order of 10 mmHg across both the lab induced and free-world datasets. All Foundation Models yield similar performance, with HiMAE and LSM-1 providing marginal improvements but no decisive advantage. The systolic task exhibits a similar profile. Across datasets, performance saturates at errors slightly around 10 mmHg, with self-supervised approaches again clustered closely together. Despite this performance, our model does achieve the lowest mean absolute error across 2 out of the 4 comparisons showing that the model design does achieve better performance under the majority of scenarios. However, despite methodological advances, the achievable error floor has yet to approach clinically useful levels \citep{mehta2024examining}.

\clearpage
\subsection{Finetuning Improves Regression Performance}

\begin{figure}
    \centering
    \includegraphics[width=0.45\linewidth]{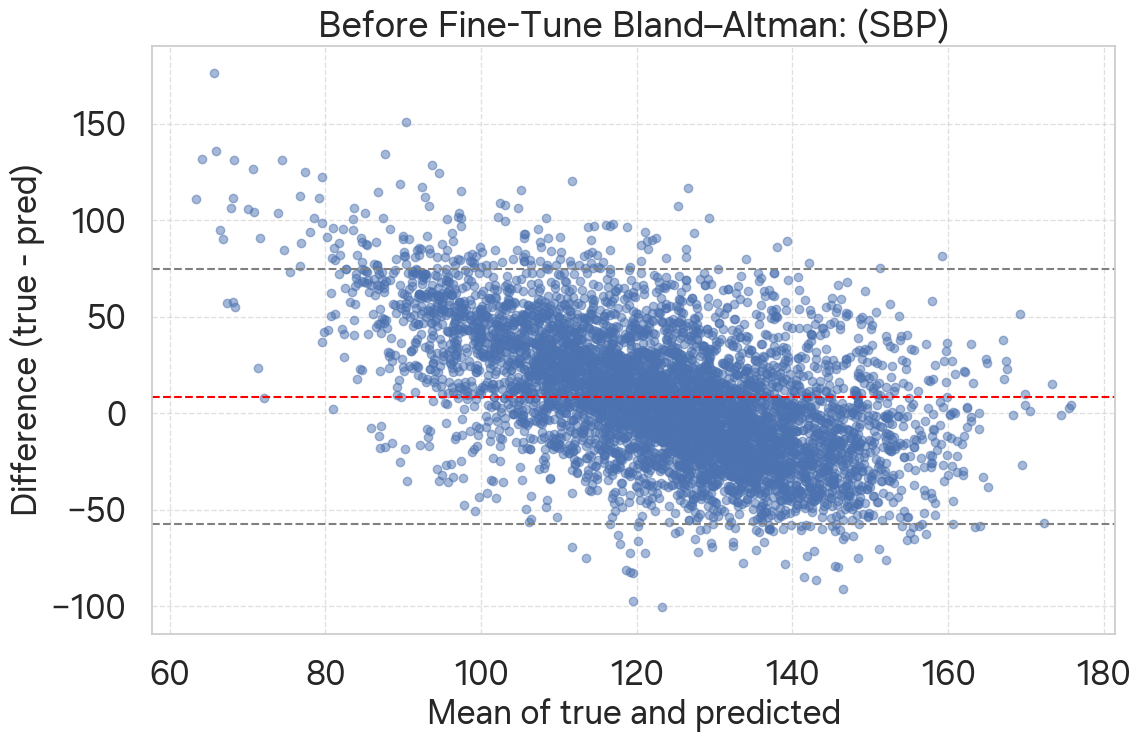}
    \includegraphics[width=0.45\linewidth]{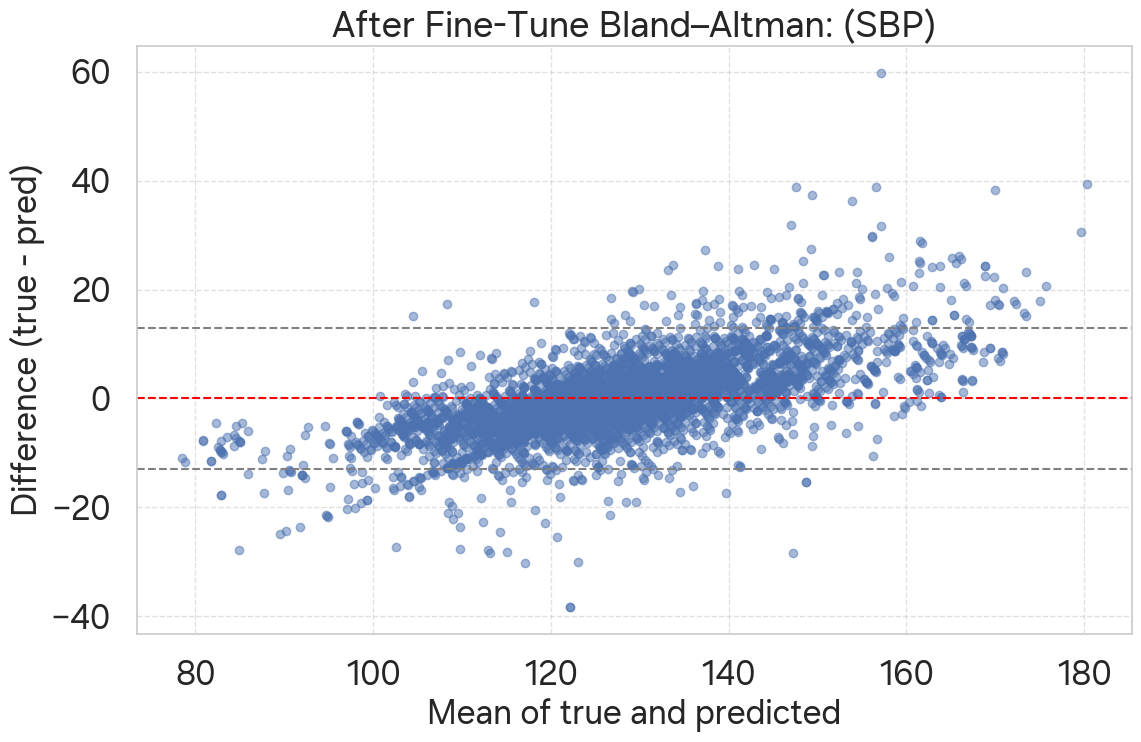}
    \includegraphics[width=0.45\linewidth]{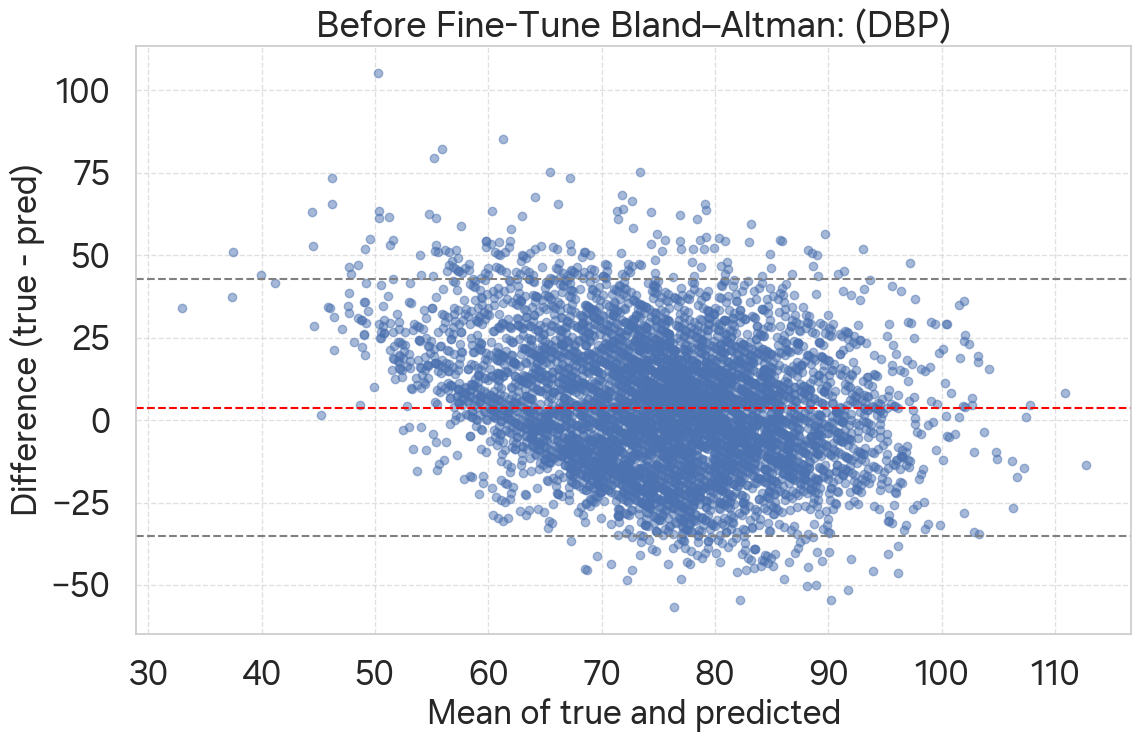}
    \includegraphics[width=0.45\linewidth]{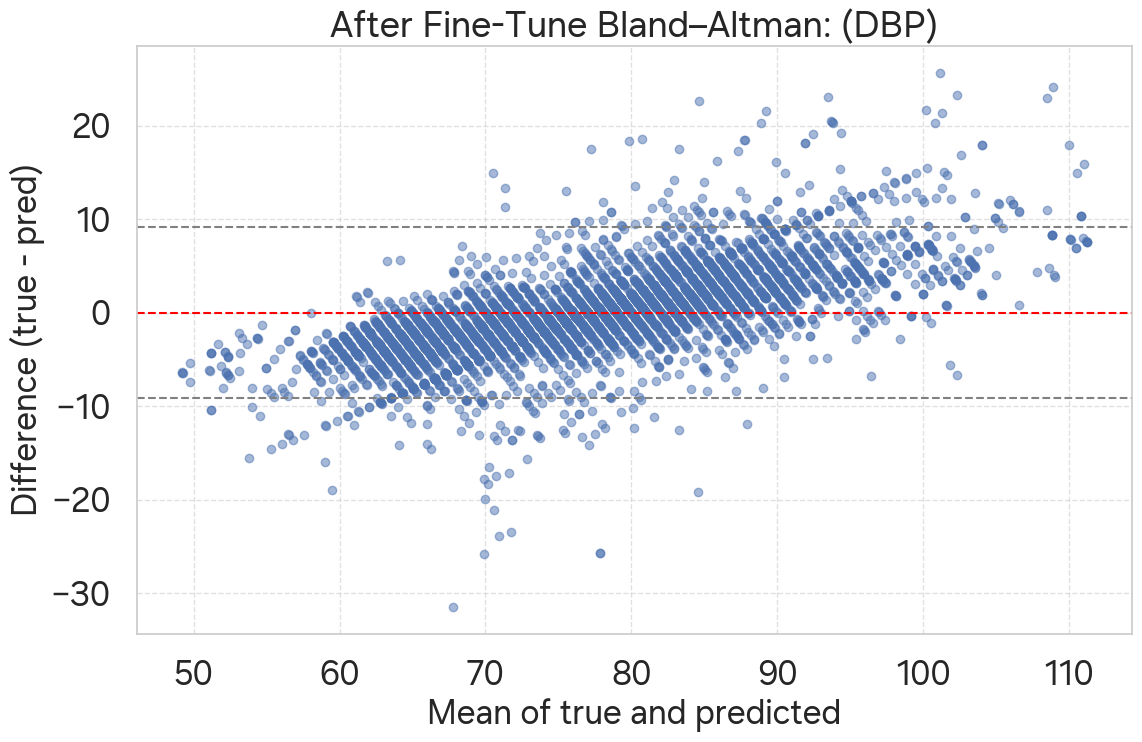}
    \caption{\textbf{Bland–Altman plot before and after fine-tuning on blood pressure regression.}  The plots illustrate the agreement between predicted and reference blood pressure values, with mean bias (solid line) and 95\% limits of agreement (dashed lines). Fine-tuning substantially reduces systematic bias and narrows the limits of agreement, indicating improved calibration and reliability of HiMAE-derived representations for regression.
}
    \label{blandaltman}
\end{figure}

Fine-tuning substantially improves the regression behavior of our blood pressure estimators, as evidenced by the Bland–Altman plots in Figure~\ref{blandaltman}. Prior to fine-tuning, both systolic and diastolic predictions exhibit large variance and systematic deviations, with wide limits of agreement and bias patterns that suggest poor calibration. After fine-tuning, the error distributions contract markedly: variance is reduced, biases approach zero, and the limits of agreement narrow considerably. These shifts indicate that fine-tuning not only enhances point prediction accuracy but also improves the overall reliability of the regression component, yielding estimates that are more clinically consistent with reference values. Despite this improvement, the model also indicates errors exceeding +/- 20mmHg which again highlight a limitation in these approaches to do well on estimating blood pressure.

\newpage
\clearpage
\subsection{TSNE plots on linear probes and fine-tuned}

\begin{figure}[h!]
    \centering

    \includegraphics[width=0.49\linewidth]{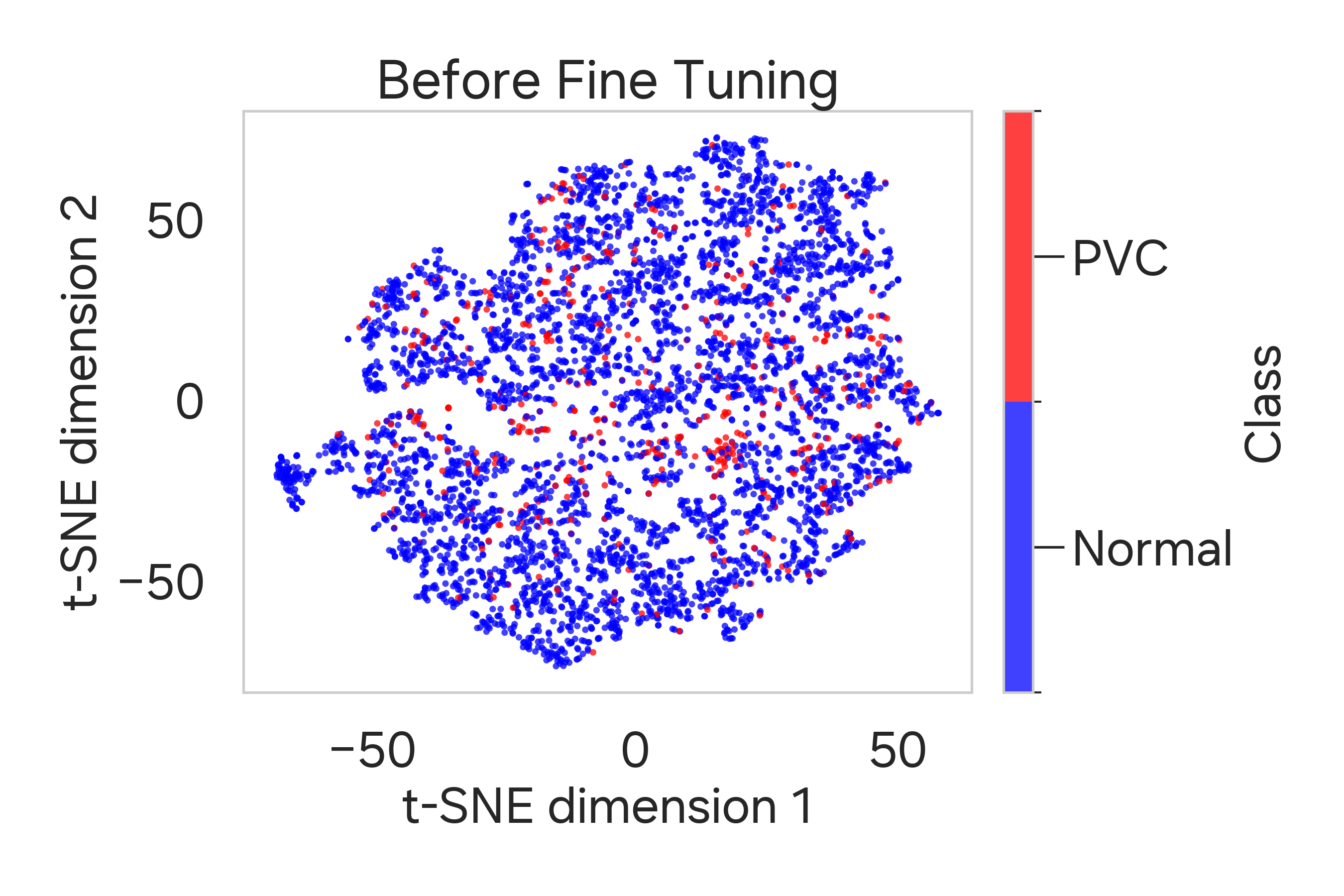}
    \includegraphics[width=0.49\linewidth]{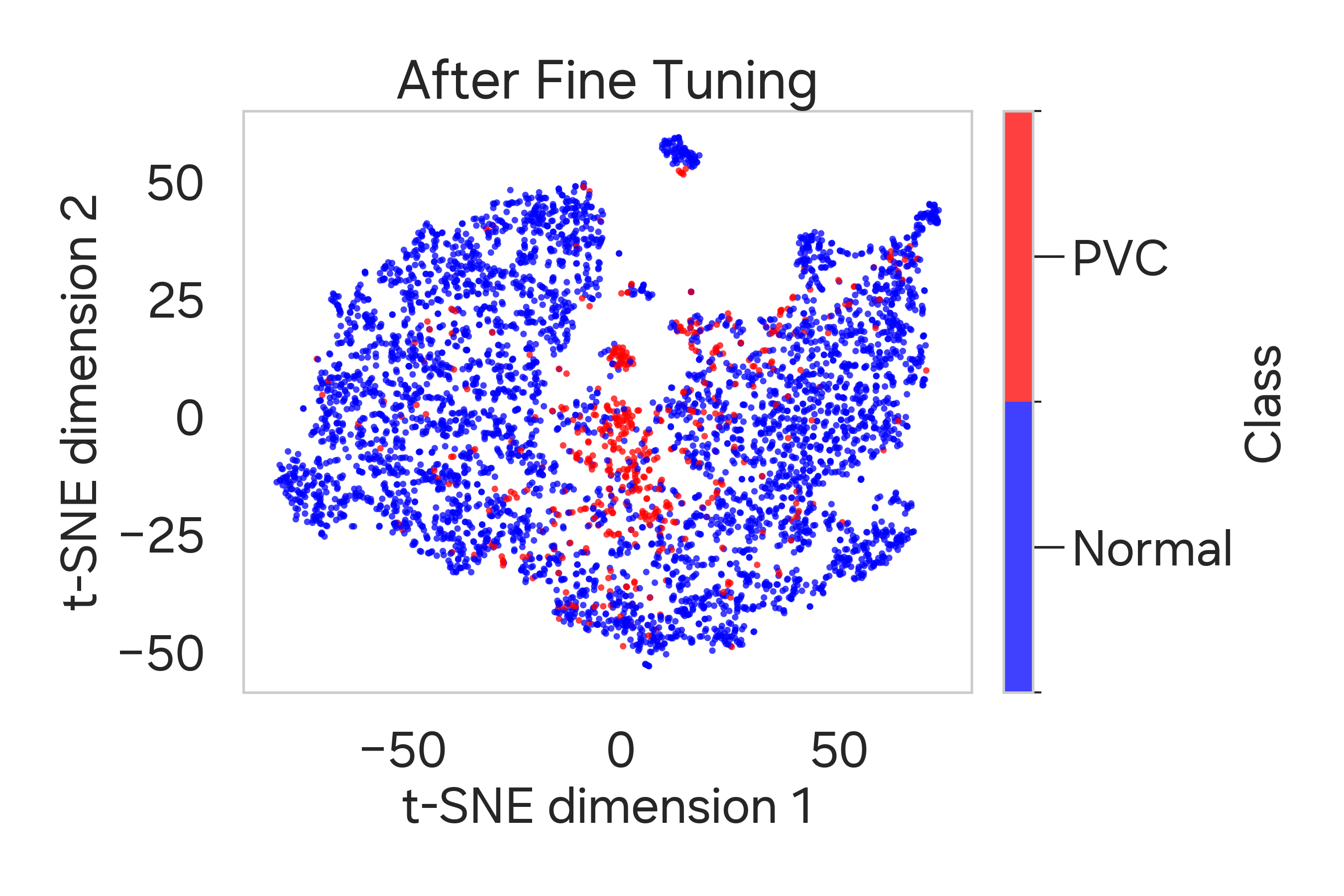}
    \includegraphics[width=0.49\linewidth]{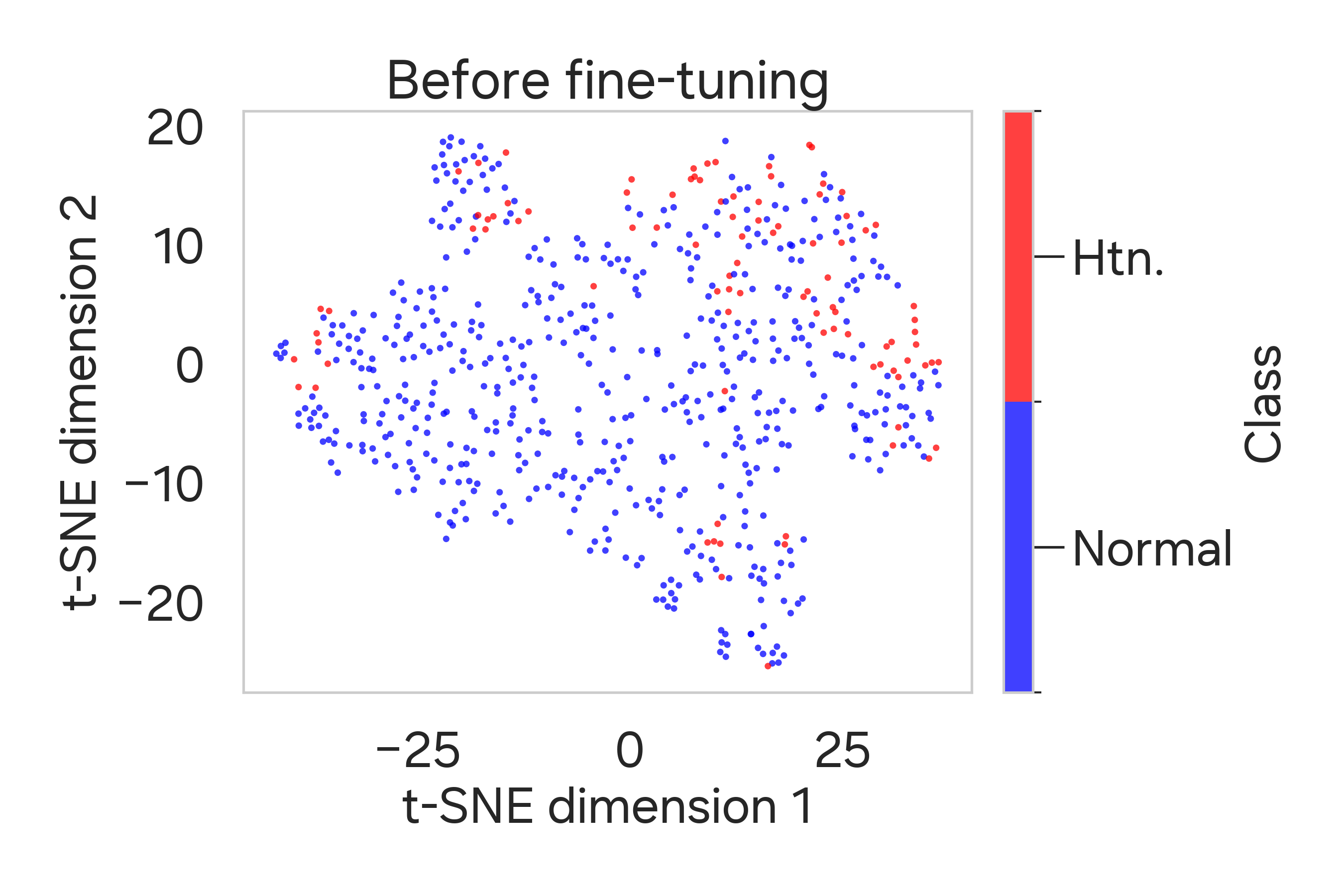}
    \includegraphics[width=0.49\linewidth]{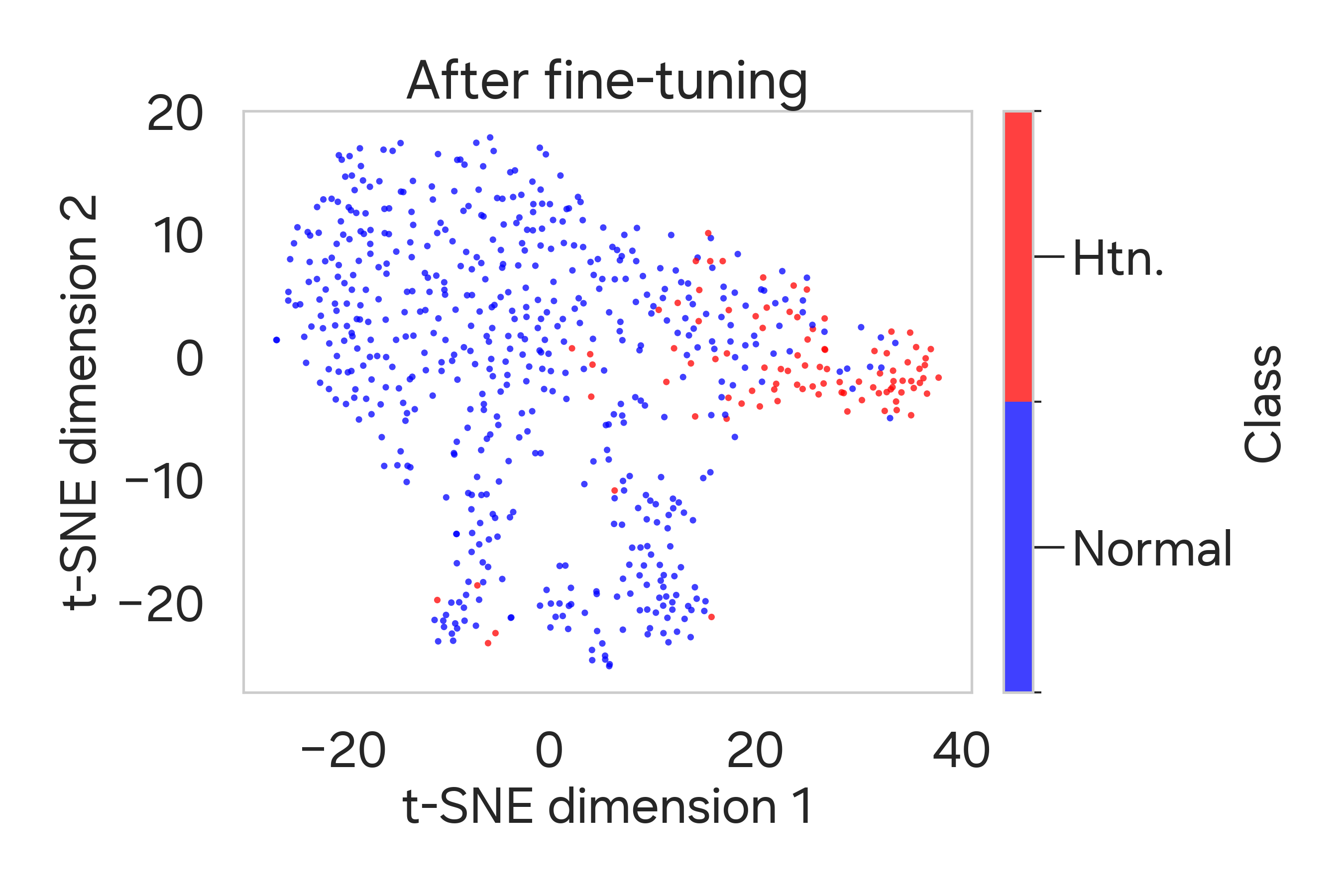}
    \caption{\textbf{t-SNE Visualization of Representations Before and After Fine-tuning.}  
Two representative tasks are shown: premature ventricular contraction (top) and hypertension detection (bottom). Each panel displays a 2D t-SNE projection of HiMAE embeddings colored by class label. Before fine-tuning, the clusters for normal and abnormal cases overlap substantially. After fine-tuning, the separation between classes becomes more pronounced, indicating that task-specific supervision sharpens decision boundaries in the learned representation space.
}
    \label{tsne}
\end{figure}

\textbf{t-SNE analysis.} Figure~\ref{tsne} visualizes embeddings using t-SNE before and after fine-tuning. Prior to fine-tuning, normal and abnormal samples form largely overlapping clusters, indicating that pretraining alone does not fully separate task-specific structure. After fine-tuning, separation between classes becomes more distinct, particularly for PVC detection, suggesting that lightweight task-specific adaptation sharpens decision boundaries while preserving the efficiency of the pretrained HiMAE representations. This confirms that HiMAE provides a strong initialization that benefits from minimal supervised refinement.

\newpage
\section{On-device Experiments}

\begin{table}[h!]
\centering
\small
\begin{tabular}{lccc}
\toprule
\textbf{Model} & \textbf{Params} & \textbf{FLOPs} & \textbf{Memory} \\
\midrule
HiMAE & 1.2M & 0.0647 gFLOPS & 4.8 MB\\
Efficient-Net & 7.8M & 0.70 gFLOPS & 31.1 MB\\
Swin-Transformer & 110.6M & 11.89 gFLOPS & 423.8 MB\\
LSM-Base   & 110.6M & 15.94 gFLOPS & 441.3 MB\\
\bottomrule
\end{tabular}
\caption{\textbf{HiMAE is lightweight and efficient:} Model size and compute cost comparison between HiMAE and LSM. FLOPs measured per forward pass on a $10$s sequence at $100$Hz.}
\label{params}
\vspace{-0.6cm}
\end{table}


\subsection{Inference Efficency}
We benchmarked the inference efficiency of our proposed HiMAE against the transformer baseline (LSM-Base), measuring three aspects: model footprint and computational complexity in terms of parameters, memory, and FLOPs per 10-second input window at 100 Hz (Table~\ref{params}); latency, defined as mean per-sample forward-pass time at batch size 1; and throughput, defined as the maximum number of samples processed per second (Table~\ref{tab:inference}). All experiments were run on a Samsung Watch Series 8. Benchmarks were run on-device, using Exynos W1000 CPUs. We also tested on a T4 GPU for potential mobile device deployment; although the T4 is a datacenter GPU, modern mobile processors like the Qualcomm Adreno 750 found on commercial phones are optimized for high-performance ML and can deliver comparable efficiency \citep{buber2018performance, wesolowski2021datacenter}, underscoring the practicality of on-device deployment.

\begin{wraptable}{r}{0.65\textwidth}
\centering
\small
\begin{tabular}{lcccc}
\toprule
\textbf{Model} & \textbf{GPU Lat.} & \textbf{GPU Thr.} & \textbf{CPU Lat.} & \textbf{CPU Thr.} \\
\midrule
HiMAE & 0.039 ms & 25.8k/s & 0.99 ms & 1.2k/s \\
Efficient-Net & 0.082 ms & 12.2k/s & 1.42 ms& 0.704k/s\\
Swin-Transformer & 0.704 ms & 1.42k/s & 2.95 ms & 0.456k/s \\
LSM-Base & 0.80 ms & 1.24k/s & 3.36 ms & 0.298k/s \\
\bottomrule
\end{tabular}
\caption{\textbf{Inference Performance:} Latency (ms per sample, batch size 2048) and throughput (samples/sec) measured over 10\,s windows.}
\label{tab:inference}
\end{wraptable}

\noindent\underline{Results}

Despite being more than two orders of magnitude smaller in parameter count, the HiMAE consistently outperforms the transformer baseline across all efficiency metrics. Between Efficient-Net \citep{tan2020efficientnetrethinkingmodelscaling}, it remains marginally better which is encouraging due to the optimizations designed in this model.

\noindent\underline{Model footprint:} 

HiMAE reduces parameters from $110$M to $0.31$M ($\sim355\times$ fewer), FLOPs from $15.94$G to $0.0647$G ($\sim246\times$ fewer), and memory from $441.3$MB to $3.6$MB ($\sim123\times$ smaller). These reductions highlight that computational savings scale with the compactness of the model, without loss of representational capacity for the task.

\noindent\underline{Latency:} 

HiMAE achieves substantially faster per-sample inference. On GPU, latency drops from $0.80$ms to $0.039$ms ($\sim20\times$ faster), while on CPU it falls from $3.93$ms to $0.99$ms ($\sim4\times$ faster). The reduction in latency follows directly from the smaller computational footprint, reflecting a consistent efficiency advantage.

\noindent\underline{Throughput:} 

These improvements translate into higher throughput across hardware. On GPU, throughput increases from $1.24$k to $25.8$k samples/s ($\sim21\times$ higher), while CPU throughput rises from $0.255$k to $1.2$k samples/s ($\sim5\times$ higher). These results confirm that computational gains extend beyond memory and FLOPs, yielding end-to-end speedups at inference time.

In summary, HiMAE achieves a favorable tradeoff between compactness and efficiency, providing lower FLOPs, smaller memory footprint, and faster inference despite its reduced model size. It also outperforms Efficient-Net B1 which was specially designed and optimized for performance and compactness giving a comparison and context to our models performance.

\end{document}